\documentclass[letterpaper]{article} 
\usepackage[]{aaai2026}  
\usepackage{times}  
\usepackage{helvet}  
\usepackage{courier}  
\usepackage[hyphens]{url}  
\usepackage{siunitx} 
\usepackage{etoolbox}
\usepackage{graphicx} 
\usepackage{natbib}  
\usepackage{caption} 
\usepackage{amsmath}

\usepackage[ruled,vlined]{algorithm2e}

\usepackage{booktabs}
\usepackage{multirow} 
\usepackage{subcaption} 

\usepackage{xcolor}
\usepackage{caption}
\usepackage{algpseudocode}
\usepackage{amssymb}
\usepackage[table]{xcolor}
\definecolor{matchcolor}{RGB}{25, 50, 180} 
\definecolor{diffcolor}{RGB}{210, 40, 40} %
\title{MoRA:  On-the-fly {Mo}lecule-aware Low-{R}ank {A}daptation Framework for LLM-based Multi-Modal Molecular Assistant}

\author{
    Tao Yin\textsuperscript{\rm 1,2},
    Xiaohong Zhang\textsuperscript{\rm 1,2},
    Jiacheng Zhang\textsuperscript{\rm 1,2},
    Li Huang\textsuperscript{\rm 1,2},
    Zhibin Zhang\textsuperscript{\rm 1,2},
    Yuansong Zeng\textsuperscript{\rm 1},
    Jin Xie\textsuperscript{\rm 1,2},
    Meng Yan\textsuperscript{\rm 1,2}
}
\affiliations{
    \textsuperscript{\rm 1}School of Big Data and Software, Chongqing University\\
    \textsuperscript{\rm 2} Key Laboratory of Dependable Service Computing in Cyber-Physical Society
}

\usepackage{bibentry}

\begin{document}

\maketitle

\begin{abstract}
Effectively integrating molecular graph structures with Large Language Models (LLMs) is a key challenge in drug discovery. Most existing multi-modal alignment methods typically process these structures by fine-tuning the LLM or adding a static adapter simultaneously. However, these approaches have two main limitations: (1) it optimizes a shared parameter space across all molecular inputs, limiting the model’s ability to capture instance-specific structural features; and (2) fine-tuning the LLM for molecular tasks can lead to catastrophic forgetting, undermining its general reasoning capabilities. In this paper, instead of static task-oriented adaptation, we propose an instance-specific parameter space alignment approach for each molecule on-the-fly. To this end, we introduce Molecule-aware Low-Rank Adaptation (MoRA) that produces a unique set of low-rank adaptation weights for each input molecular graph. These weights are then dynamically injected into a frozen LLM, allowing the model to adapt its reasoning to the structure of each molecular input, while preserving the LLM’s core knowledge. Extensive experiments demonstrate that on key molecular tasks, such as chemical reaction prediction and molecular captioning, MoRA's instance-specific dynamic adaptation outperforms statically adapted baselines, including a 14.1\% relative improvement in reaction prediction exact match and a 22\% reduction in error for quantum property prediction. The code is available at https://github.com/jk-sounds/MoRA.

\end{abstract}


\section{Introduction}
Molecular machine learning has become an important area of research due to its broad applications in both scientific and industrial contexts. Recently, the development of Large Language Models (LLMs)\cite{achiam2023gpt,touvron2023llama} has introduced promising opportunities for advancing molecular science, particularly in fields such as drug discovery and materials design\cite{chakraborty2025ai, chakraborty2023artificial}.

\begin{figure}
    \centering
    \includegraphics[width=1.\linewidth]{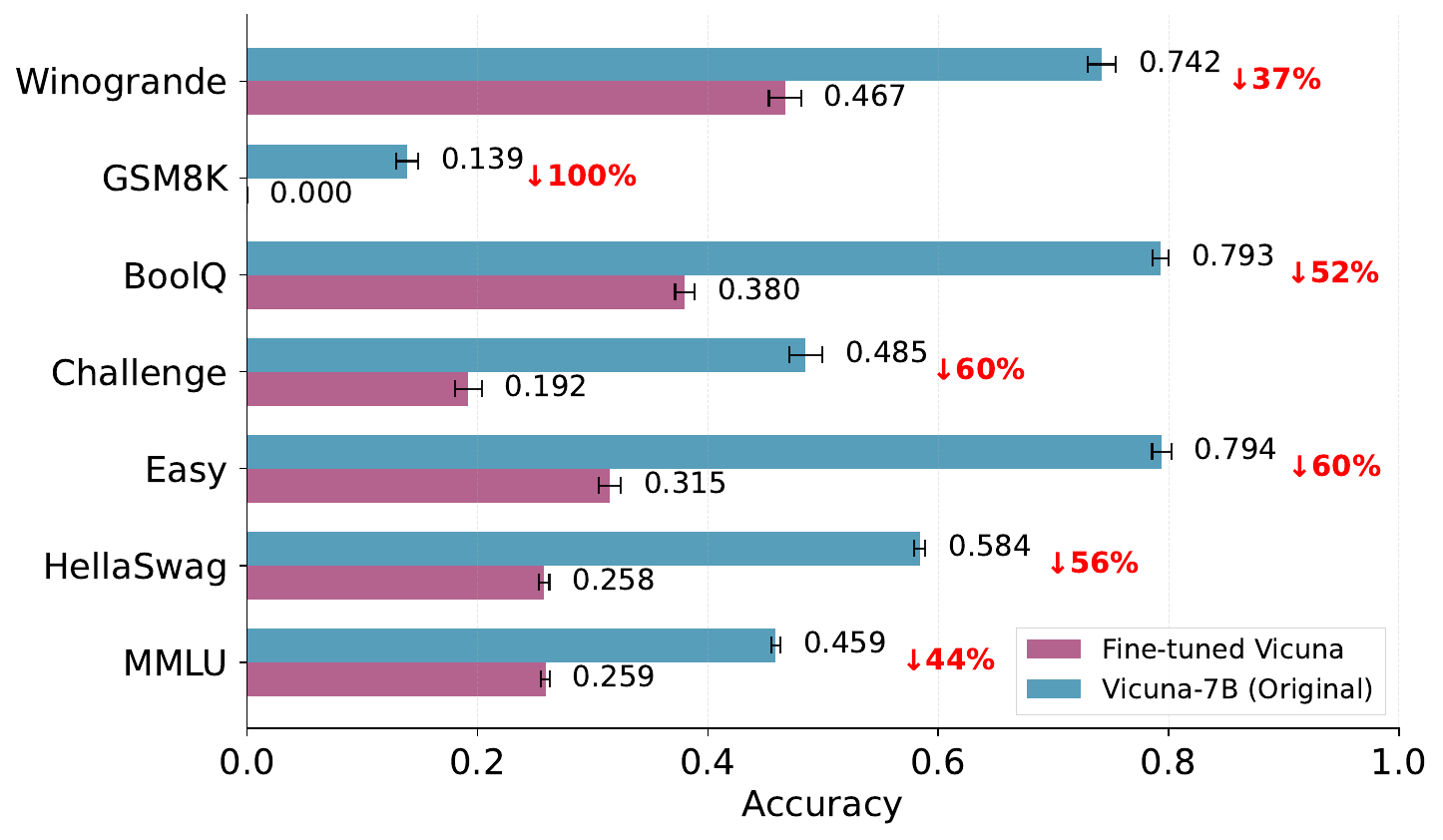}
    \caption{Catastrophic forgetting in vicuna-7B after domain-specific LoRA fine-tuning on Mol-Instruction datasets. Original model performance (blue) versus fine-tuned model (red) across general benchmarks.}
    \label{fig:performance_comparison}
\end{figure}
\begin{figure*}
    \centering
    \includegraphics[width=1.0\linewidth]{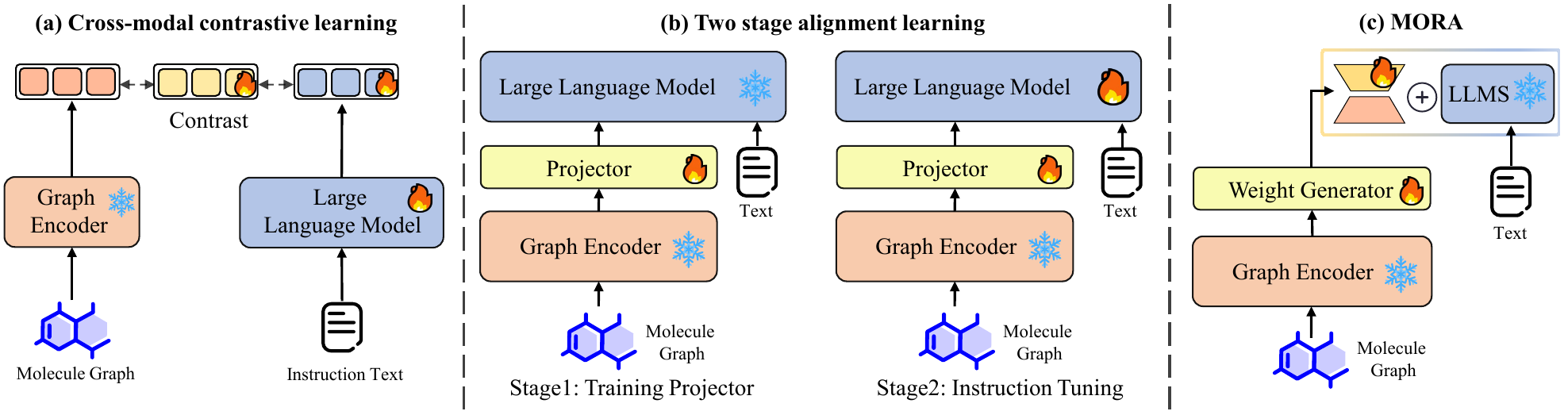}
    \caption{Comparison of molecular multimodal language modeling.}
    \label{fig:Comparison}
\end{figure*}
Traditional molecular language modeling has primarily relied on one-dimensional sequence representations~\cite{edwards2022translation,zeng2022deep,irwin2022chemformer}, such as SMILES~\cite{weininger1988smiles} and SELFIES~\cite{krenn2022selfies,krenn2020self}. However, these representations fail to adequately capture the crucial 2D molecular graph structures, limiting the model's performance on complex tasks~\cite{wang2022molecular,su2022molecular}. Consequently, recent studies have begun integrating 2D molecular graphs as a core modality into model architectures ~\cite{fang2023mol,luo2023molfm,cao2025instructmol}. Another key development is multi-task instruction fine-tuning~\cite{liu2023multi,park2024llamo}, which aims to equip LLMs with the ability to follow complex molecular-related instructions across diverse tasks. Currently, the methods for incorporating molecular graph information into LLMs mainly follow two mainstream paradigms. The first is Decoupled Contrastive Alignment  (Figure~\ref{fig:Comparison}a)~\cite{tang2024mollm, text2mol,liu2023multi,su2022molecular,luo2023molfm,liu2025text}, which aligns the global representations of molecules and text in independent latent spaces. The second is a Two-Stage Input-space projection alignment (Figure~\ref{fig:Comparison}(b))~\cite{cao2025instructmol,lee2025mol,park2024llamo,fang2024moltc,kim2025mol,le2024molx}, which directly projects molecular graph features as input embeddings of LLMs.

These paradigms usually fine-tune LLMs using parameter-efficient fine-tuning (PEFT) techniques represented by low-rank adaptation (LoRA) ~\cite{hu2022lora}, which effectively adapts LLMs to the overall distribution of specific molecular tasks ~\cite{li2025large,lee2025mol,cao2025instructmol,kim2025mol}. However, molecules have complex structures, and a single static weight is difficult to fully express the subtle structural features of each individual molecule.
In addition, these paradigms all face a fundamental balance between domain specialization and model generality. As shown in Figure~\ref {fig:performance_comparison}, the parameter-efficient fine-tuning process also inevitably compromises the underlying general knowledge of the model. This catastrophic forgetting results in the LLM sacrificing its powerful general reasoning ability while gaining specific molecular capabilities. Consequently, the model becomes a "narrow expert" and fails to meet the broader demands of a molecular multimodal assistant, which is expected to handle specialized tasks while also supporting dialogue, instruction following, and scientific reasoning.

To alleviate this fundamental conflict, we advocate a shift from static task-oriented fine-tuning to dynamic instance-specific adaptation for molecular modalities. We introduce MoRA, a novel multi-modal framework that imbues an LLM with molecular understanding capabilities without altering its core parameters. Instead of learning a static task-oriented LoRA adapter, MoRA employs a Molecule-Aware Weight Generator (MAW-Gen) that dynamically produces low-rank, instance-specific adaptation weights ($\Delta W$) based on the input molecular graph. These instance-specific weights are then injected into the frozen LLM, modulating its computational pathways to instill structure-awareness within a single inference pass. Unlike conventional approaches, MoRA (Figure~\ref{fig:Comparison}c) skips explicit modality alignment and avoids modifying the base LLM’s parameters. During inference, the generated $\Delta W$ is directly injected into the LLM’s key modules (e.g., self-attention and feed-forward layers), enabling structure-aware, non-destructive computation. This design empowers the LLM to perceive molecular context without compromising its general knowledge or task-agnostic reasoning ability. We summarize our main contributions as follows:

\begin{figure*}
    \centering
    \includegraphics[width=1.\linewidth]{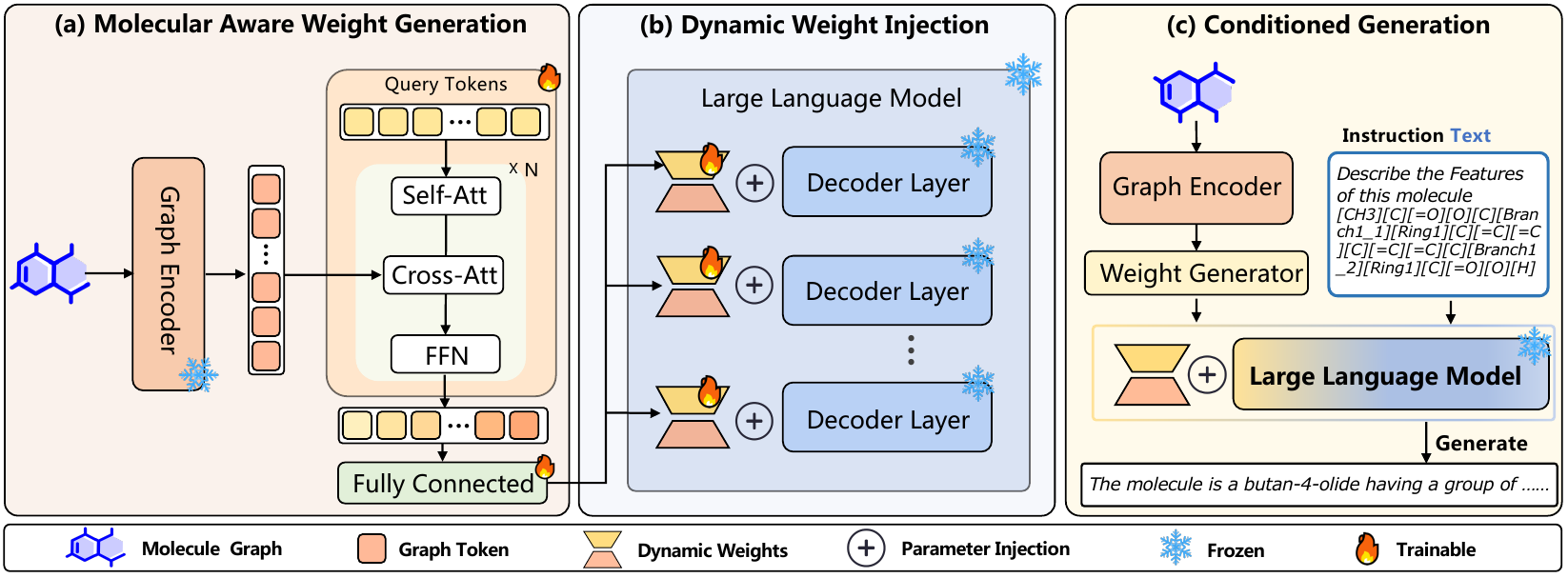}
    \caption{The overall architecture of our proposed MoRA framework. It comprises a graph encoder, a Molecule-Aware Weight Generator (MAW-Gen), and a frozen Large Language Model. MoRA first encodes an input 2D molecular graph with the graph encoder, then converts the encoded graph into a set of low-rank weight matrices via the MAW-Gen. These weights are subsequently injected into the parameters of the frozen LLM. Finally, the instruction-following response is generated by the LLM, conditioned on both the text prompt and its newly injected, molecule-aware weights.}
    \label{fig:model}
\end{figure*}

\begin{itemize}
    \item We introduce MoRA, a novel multi-modal adaptation framework that integrates molecular graph information into a frozen LLM on-the-fly, enabling strong performance on molecular tasks while mitigating the catastrophic forgetting.
    \item We design and implement the Molecule-Aware Weight Generator, which translates molecular graph structures into instance-specific, low-rank adaptation weights for the LLM.
    \item Extensive experiments demonstrate that Mora achieves state-of-the-art performance on a wide range of comprehension and generation tasks, while also providing new perspectives for molecule generation.
\end{itemize}

\section{Related Work}
\subsection{Molecular Multimodal Learning}
Integrating molecular structures with Large Language Models (LLMs) has become a prominent research direction in the pursuit of developing foundational models for science \cite{cao2025instructmol,kim2025mol,lee2025mol}. Prevailing approaches~\cite{edwards2022translation,zeng2022deep} have primarily focused on aligning representations from textual descriptions with those from molecular graphs. Early methods, inspired by successes in vision-language pre-training \cite{dai2023instructblip,radford2021learning,singh2022flava,wang2021simvlm}, utilized contrastive learning to align global representations of molecular graphs and their textual descriptions \cite{edwards2022translation,irwin2022chemformer}. However, this global alignment strategy, while effective for retrieval tasks, fails to provide the fine-grained, conditional control necessary for generative tasks such as molecule captioning or reaction prediction~\cite{tran2025xmolcap,gu2021domain,chen2025reactgpt}. Consequently, the dominant paradigm shifted towards methods that connect a modality-specific encoder to the LLM via a trainable projector module, a strategy popularized by vision-language models like BLIP-2 \cite{li2023blip} and LLaVA \cite{liu2023visual}. These approaches typically employ a two-stage training process: an initial alignment phase where the projector maps encoded features into the LLM's input space, followed by end-to-end instruction fine-tuning of the entire system. This methodology has become the de facto standard for state-of-the-art multimodal molecular LLMs \cite{cao2025instructmol,le2024molx,lee2025mol}. Unlike methods that use static adapters for contrastive and input-space alignment, inspired by HyperNetworks \citep{ha2016hypernetworks}, MoRA dynamically generates unique parameter adapters for each molecule to directly modulate a frozen LLM's internal computational pathways, enabling deeper, fine-grained molecular understanding without sacrificing its general reasoning capabilities. Specifically, our Molecule-Aware Weight Generator (MAW-Gen) acts as a graph-conditioned hypernetwork, creating instance-specific low-rank weights on-the-fly to dynamically tailor the computational paths of a frozen LLM for each molecule. 
\subsection{Parameter-Efficient Fine-Tuning}
The high computational and memory costs of large language models~\cite{touvron2023llama} render full-parameter fine-tuning impractical for most applications. This challenge has catalyzed the development of Parameter-Efficient Fine-Tuning (PEFT) methods, which adapt pre-trained models by updating only a small subset of their parameters~\cite{han2024parameter}.
Prominent PEFT strategies include additive and prefix-based methods. Low-Rank Adaptation ({LoRA}) \cite{hu2022lora} represents an influential paradigm operating directly in the parameter space. LoRA is predicated on the hypothesis that the weight update matrix during adaptation exhibits a low intrinsic rank. Consequently, it freezes the pre-trained weight matrix $\mathbf{W}_0$ and injects a parallel, trainable low-rank decomposition $\Delta\mathbf{W} = \mathbf{B}\mathbf{A}$. During fine-tuning, only the low-rank matrices $\mathbf{A}$ and $\mathbf{B}$ are optimized. 

\section{Method}
We introduce MoRA, a novel framework for multi-modal molecular instruction-following that operates by directly modulating the parameter space of a Large Language Model (LLM). Unlike conventional approaches that align modalities in the input or feature space, MoRA generates molecule-specific low-rank weight adaptations for a frozen LLM, allowing the LLM to perform structure-aware reasoning conditioned on text prompts. For tasks without molecular graphs, our method switches to utilizing the frozen backbone to infer textual prompts,  preserving the full capabilities of the pre-trained model.

\subsection{Problem Formulation}
To build a versatile molecular assistant, we focus on the task of multi-modal instruction-following. The goal is to develop a model that understands natural language instructions alongside molecular structural data, enabling it to generate accurate textual answers for a wide array of chemical tasks. Formally, each data instance is a triplet $(I, \mathcal{G}, A)$, where $I$ is a textual instruction, $A$ is the target textual answer, and $\mathcal{G}$ represents the associated molecular information. While a molecule is a multi-modal entity, we primarily focus on its 2D structural graph, denoted as $\mathcal{G} = (\mathcal{V}, \mathcal{E})$, which captures atomic composition and connectivity. The answer $A$ can range from a numerical value or a descriptive paragraph to a new molecular sequence for generative tasks like retrosynthesis. Our objective is to learn a single model, parameterized by a set of trainable parameters $\theta_{\text{train}}$, that approximates the conditional probability $P(A | I, \mathcal{G})$. The model's parameters are optimized by maximizing the log-likelihood over a multi-task instruction-tuning dataset $\mathcal{D}$:
\begin{equation}
\max_{\theta_{\text{train}}} \sum_{(I, \mathcal{G}, A) \in \mathcal{D}} \log P(A | I, \mathcal{G}; \theta_{\text{train}})
\label{eq:objective}
\end{equation}
The specific composition of these trainable parameters, which exclusively belong to our proposed generator module while all base model weights remain frozen, is detailed in Section \textit{Training Method}.

\subsection{MoRA Architecture}
The MoRA framework, depicted in Figure~\ref{fig:model}, is composed of three primary components: (1) a Molecule-Aware Weight Generator (MAW-Gen) that translates graph embeddings into LLM parameter updates; (2) a Dynamic Weights Injection mechanism where the generated weights augment each frozen decoder layer parametrically; and (3) a Conditioned Generation stage in which the structurally-augmented LLM serves as the reasoning backbone to produce the final output.

\subsection{Graph Encoder}

To perceive molecular structures, we employ a Graph Neural Network (GNN), denoted as \(g(\cdot)\), to process the input molecular graph \(\mathcal{G} = (\mathcal{V}, \mathcal{E})\). The GNN operates via a message-passing scheme over \(L\) layers. At each layer \(l\), this process involves two steps. First, an aggregated message \(\mathbf{m}_{\mathcal{N}(v)}^{(l)}\) is computed from the representations of neighboring nodes:
\begin{equation}
    \mathbf{m}_{\mathcal{N}(v)}^{(l)} = \mathbf{AGGREGATE}^{(l)} \left( \{ \mathbf{h}_u^{(l-1)} \}_{u \in \mathcal{N}(v)} \right).
\end{equation}
Then, the representation of node \(v\), \(\mathbf{h}_v^{(l)}\), is updated by combining its previous state with this aggregated message:
\begin{equation}
    \mathbf{h}_v^{(l)} = \mathbf{UPDATE}^{(l)} \left( \mathbf{h}_v^{(l-1)}, \mathbf{m}_{\mathcal{N}(v)}^{(l)} \right),
\end{equation}
where \(\mathbf{h}_v^{(0)}\) is initialized with the input atom features.
After \(L\) iterations, each final node embedding \(\mathbf{h}_v^{(L)}\) effectively captures the rich chemical environment within its \(L\)-hop receptive field. 
The complete set of node embeddings, $\mathbf{H}_\mathcal{G} = \{\mathbf{h}_v^{(L)} \mid v \in \mathcal{V}\}$, which encapsulates the molecule's structural topology, is then passed to the subsequent MAW-Gen module.

\subsection{Molecule-Aware Weight Generator}
The Molecule-Aware Weight Generator (MAW-Gen) is the core of the MoRA framework, responsible for bridging the modality gap between the graph encoder's feature space and the LLM's parameter space. It transforms the set of graph node embeddings $\mathbf{H}_\mathcal{G}$ into a series of low-rank matrices that modulate the LLM's weights. This process consists of two stages: a cross-attention distillation head and a low-rank parameter projection head.
\subsubsection{Cross-Attention Distillation Head.} Inspired by query-based mechanisms in vision-language models, we distill the node-level graph features into a set of compact representations tailored for parameter generation. We introduce a small set of $k$ learnable molecular queries, $\mathbf{Q}_{\text{learn}} \in \mathbb{R}^{k \times d_{\text{model}}}$. These queries are processed by a stack of transformer decoder layers. Within each layer, the queries first attend to each other via self-attention, allowing them to specialize. Subsequently, a cross-attention mechanism enables the queries to attend to the set of node embeddings $\mathbf{H}_\mathcal{G}$, aggregating relevant structural and chemical information. The output of this process is a set of $k$ context-aware query vectors, $\mathbf{Q}_{\text{out}} \in \mathbb{R}^{k \times d_{\text{model}}}$, which have effectively distilled the molecular structure into a format suitable for the next stage.
\subsubsection{Low-Rank Parameter Projection.}
Generating full-rank weight matrices for the LLM is infeasible. We thus adopt a low-rank factorization strategy inspired by LoRA~\cite{hu2022lora}. For each target component $c$ in the LLM (e.g., the query projection matrix $\mathbf{W}_q$), the molecule-specific weight update, denoted as $\mathbf{W}_c \in \mathbb{R}^{d_{\text{llm}} \times d_{\text{llm}}}$, is factorized into two smaller matrices, $\Delta\mathbf{A}_c \in \mathbb{R}^{d_{\text{llm}} \times r}$ and $\Delta\mathbf{B}_c \in \mathbb{R}^{r \times d_{\text{llm}}}$, where the rank $r \ll d_{\text{llm}}$. We establish a one-to-one mapping between the $k$ distilled query vectors in $\mathbf{Q}_{\text{out}}$ and the set of all target components. For a given component $c$, we select its corresponding query vector $\mathbf{q}_c$ from $\mathbf{Q}_{\text{out}}$. The instance-specific matrix $\Delta\mathbf{A}_c$ is generated by projecting $\mathbf{q}_c$ using a fully-connected layer, $\mathbf{W}_{\text{FC}}$, and then reshaping the result. The matrix $\Delta\mathbf{B}_c$ is realized as a shared, learnable linear projector, $\mathbf{W}_{\text{proj}}$, which is constant for all generated weights and is analogous to the up-projection matrix in standard LoRA. The complete weight update for a specific component $c$, which we denote generically as $\mathbf{W}_{\text{mol}}$, is thus formulated as:
\begin{equation}
    \mathbf{W}_{\text{mol}} = \Delta\mathbf{A}_c \cdot \Delta\mathbf{B}_c = R(\mathbf{q}_c \mathbf{W}_{\text{FC}}) \mathbf{W}_{\text{proj}},
\end{equation}
where $R(\cdot)$ is an operator that reshapes its input vector from $\mathbb{R}^{d_{\text{llm}} \cdot r}$ into a matrix in $\mathbb{R}^{d_{\text{llm}} \times r}$. The layers $\mathbf{W}_{\text{FC}} \in \mathbb{R}^{d_{\text{model}} \times (d_{\text{llm}} \cdot r)}$ and $\mathbf{W}_{\text{proj}} \in \mathbb{R}^{r \times d_{\text{llm}}}$ are learnable parameters. This factorization imposes a strong regularization by decoupling the generation process: the MAW-Gen only maps the query $\mathbf{q}_c$ into a low-rank bottleneck via $\mathbf{W}_{\text{FC}}$, while the shared $\mathbf{W}_{\text{proj}}$ handles the final projection.

\begin{table*}[t]
\centering
\label{tab:main_results}
\begin{tabular}{l ccccccc}
\toprule
\textbf{Model} & \textbf{Exact} $\uparrow$ & \textbf{BLEU} $\uparrow$ & \textbf{Levenshtein} $\downarrow$ & \textbf{RDK FTS} $\uparrow$ & \textbf{MACCS FTS} $\uparrow$ & \textbf{Morgan FTS} $\uparrow$ & \textbf{Validity} $\uparrow$ \\
\midrule
\rowcolor{gray!20}\multicolumn{8}{l}{\textit{Forward Reaction Prediction}} \\
\midrule
Alpaca†          & 0.000  & 0.065 & 41.989 & 0.004 & 0.024 & 0.008 & 0.138 \\
Baize†          & 0.000  & 0.044 & 41.500 & 0.004 & 0.025 & 0.009 & 0.097 \\
Vicuna†         & 0.000  & 0.057 & 41.690 & 0.007 & 0.016 & 0.006 & 0.138 \\
LLama†         & 0.000  & 0.020 & 42.002 & 0.001 & 0.002 & 0.001 & 0.039 \\
Text+Chem T5    & 0.239  & 0.782 & 20.413 & 0.705 & 0.789 & 0.652 & 0.762 \\
Mol-Instruction & 0.045  & 0.654 & 27.262 & 0.313 & 0.509 & 0.262 & 1.000 \\
PRESTO          & 0.355  & 0.836 & 10.647 & 0.646 & 0.726 & 0.624 & 0.973 \\
HIGHT            & 0.293  & 0.935 & 16.687 & 0.774 & 0.618 & 0.566 & 1.000 \\
InstructMol    & 0.536  & 0.967 & 10.851 & 0.776 & 0.878 & 0.741 & 1.000 \\
LLaMo          & 0.584  & 0.894 & \textbf{6.162} & {0.857} & 0.918 & {0.841} & 0.938 \\
UniMoT          & 0.611 & 0.980 & 8.297   & 0.836 & 0.911 & 0.807 & 1.000 \\
\hline

\textbf{MoRA} & \textbf{0.697} & \textbf{0.985} & {6.780}  & \textbf{0.875} & \textbf{0.923} & \textbf{0.843} & \textbf{1.000} \\
\midrule
\rowcolor{gray!20}\multicolumn{8}{l}{\textit{Retrosynthesis}} \\
\midrule
Alpaca†         & 0.000  & 0.063 & 46.915 & 0.005 & 0.023 & 0.007 & 0.160 \\
Baize†          & 0.000  & 0.095 & 44.714 & 0.025 & 0.050 & 0.023 & 0.112 \\
Vicuna†         & 0.000  & 0.057 & 46.877 & 0.025 & 0.030 & 0.021 & 0.017 \\
LLama†          & 0.000  & 0.283 & 53.510 & 0.136 & 0.294 & 0.106 & 1.000 \\
Text+Chem T5    & 0.141  & 0.765 & 24.043 & 0.685 & 0.765 & 0.585 & 0.698 \\
Mol-Instruction & 0.009  & 0.705 & 31.227 & 0.283 & 0.487 & 0.230 & 1.000 \\
PRESTO           & 0.275  & 0.902 & 14.433 & 0.655 & 0.747 & 0.619 & 0.980 \\
HIGHT            & 0.202  & 0.914 & 20.194 & 0.772 & 0.623 & 0.577 & 0.999 \\
InstructMol     & 0.407  & 0.941 & 13.967 & 0.753 & 0.852 & 0.714 & 1.000 \\
LLaMo            & 0.341  & 0.830 & 12.263 & 0.793 & 0.868 & 0.750 & 0.954 \\
UniMoT     & 0.478 & \textbf{0.974} & 11.634 & 0.810 & 0.909 & 0.771 & 1.000 \\
\hline
\textbf{MoRA} & \textbf{0.530} & {0.960} & \textbf{10.720} & \textbf{0.820} & \textbf{0.918} & \textbf{0.782} & \textbf{1.000} \\
\bottomrule
\end{tabular}
\caption{Performance comparison on forward reaction prediction and retrosynthesis tasks. The arrows (\(\uparrow, \downarrow\)) indicate whether higher or lower values are better. †: few-shot ICL results from~\cite{fang2023mol}.}
\label{tab:main_results}
\end{table*}

\subsection{Dynamic Parameter Injection}
The molecule-aware weight update, $\mathbf{W}_{\text{mol}}$, generated as defined in the preceding section, is dynamically injected into the frozen LLM's parameters. This process creates a temporary, molecule-specific computational path by augmenting the original weights. For a given pre-trained weight matrix $\mathbf{W}$ (e.g., $\mathbf{W}_q, \mathbf{W}_k, \mathbf{W}_v, \mathbf{W}_o$ in self-attention, or MLP projections), the modified weight matrix $\hat{\mathbf{W}}$ used in the forward pass is formulated as a direct summation:
\begin{equation}
    \hat{\mathbf{W}} = \mathbf{W} + \mathbf{W}_{\text{mol}}.
\end{equation}
This additive modification is applied across designated layers and components of the LLM. By directly absorbing these molecule-aware updates into its parameter space, the LLM can perceive complex chemical topologies. Crucially, because the base weights $\mathbf{W}$ are never permanently altered and the update $\mathbf{W}_{\text{mol}}$ is transient for each instance, this process enables deep, structure-aware reasoning without  degrading the model's foundational linguistic knowledge.

\subsection{Training Method}
\label{sec:training_method}
During MoRA training, the parameters of the backbone LLM ($\theta$) and GNN graph encoder ($\phi$) are kept entirely frozen. Only the parameters of the MAW-Gen ($\psi$) are updated.
For each instance $(\mathcal{G}, I, A)$ in the training set $\mathcal{D}$, the model's objective is to minimize the standard autoregressive cross-entropy loss for the target sequence $A = (a_1, \dots, a_T)$:
\begin{equation}
    \mathcal{L}(\psi) = - \sum_{t=1}^{T} \log P(a_t \mid a_{<t}, I, \mathcal{G}; \psi, \phi, \theta)
\end{equation}
This objective allows the MAW-Gen to directly learn the optimal mapping from graph structures to effective parameter modulations that guide the LLM toward generating the correct response.

\section{Experiments}
In this section, we first conduct extensive experiments on various downstream molecular tasks to demonstrate the effectiveness of our proposed approach. Afterwards, we study the performance comparison between task-oriented static model fine-tuning and instance-specific dynamic adaptation. Finally, we analyze the impact of key design choices, such as injection strategy and depth of weight generator.
\subsection{Experiment Setup}
\paragraph{Benchmarks.}
To assess MoRA's efficacy, we evaluate it on four diverse molecular reasoning tasks: (1) forward reaction prediction, (2) retrosynthesis, (3) molecule description generation, and (4) property prediction. 

\paragraph{Baselines.}
To evaluate MoRA, we benchmark it against three categories of baseline models. (1) {General LLMs}: We compare against prominent LLMs that operate solely on SMILES representations, including LLaMA2-7B~\cite{touvron2023llama}, Vicuna-7B~\cite{chiang2023vicuna}, Alpaca~\cite{taori2023stanford}, Baize~\cite{xu2023baize}. (2) Molecule-specialized LLM such as Text+Chem T5~\cite{christofidellis2023unifying}.  (3) Input-Space alignment models such as Mol-Instruction~\cite{fang2023mol}, InstructMol~\cite{cao2025instructmol},PRESTO~\cite{cao2024presto}, HIGHT~\cite{chen2024hight}, LLaMo~\cite{park2024llamo}, 3D-MoLM~\cite{li2024towards}, and UniMoT~\cite{christofidellis2023unifying}.

\paragraph{Evaluation Metrics.}
We evaluate MoRA's performance across a diverse range of molecular tasks using established, task-specific metrics. For tasks involving molecular structure generation, such as forward reaction prediction and retrosynthesis, evaluation is performed on the generated SMILES strings. We report the {Exact Match (EM)} rate for chemically valid outputs, verified using RDKit. To provide a more nuanced assessment that also rewards functional similarity, we use the {Tanimoto Similarity} based on Morgan fingerprints. For molecule description generation tasks, we assess performance using standard n-gram-based metrics: {BLEU}, {ROUGE-L}, and {METEOR}. For quantitative molecular property prediction, we report the {Mean Absolute Error (MAE)} between the predicted and ground-truth values.

\paragraph{Implementation Details.}
MoRA is instantiated with the Vicuna-7B model, whose original parameters are kept frozen throughout training. We employ a GNN-based graph encoder and our proposed Molecule-Aware Weight Generator (MAW-Gen) as the only trainable components. The MAW-Gen architecture is configured with \(N=8\) decoder blocks and \(k=4\) learnable molecular queries. The rank for the generated low-rank matrices is set to \(r=64\). These molecule-specific weight updates are injected into the query, key, value, and output projection matrices ($\mathbf{W}_q, \mathbf{W}_k, \mathbf{W}_v, \mathbf{W}_o$) within each self-attention layer of the LLM. Consistent with established practices for LoRA-based methods~\cite{hu2022lora}, the linear projection head in MAW-Gen that generates the $\Delta\mathbf{A}$ matrix is initialized with zeros to ensure training stability. For optimization, we use the AdamW optimizer to minimize the standard cross-entropy loss for text generation, and the learning rate is set to 2e-5. All experiments are conducted using 8 NVIDIA A800 GPUs with a global batch size of 96.

\subsection{Main Results}
To evaluate the effectiveness of our MoRA framework, we conduct a comprehensive comparison against a suite of baseline models on four challenging chemical reasoning tasks: forward reaction prediction, retrosynthesis, Molecule Description Generation, and Property Prediction.

\paragraph{Chemical Reaction Prediction.}
As summarized in Table~\ref{tab:main_results}, our model, {MoRA}, establishes a new state of the art on both forward reaction prediction and retrosynthesis. Specifically, in forward reaction prediction, {MoRA} achieves an Exact Match (EM) accuracy of {0.697}, outperforming the previous leading model, UniMoT, by a relative margin of 14.1\%. This performance advantage is even more pronounced in the challenging task of retrosynthesis, where {MoRA} attains an EM score of {0.530}, representing a significant 10.9\% relative improvement over UniMoT. Notably, {MoRA}'s superiority extends across most key metrics; it consistently achieves top scores in various structural similarity evaluations (RDK FTS, MACCS FTS, and Morgan FTS). While UniMoT registers a slightly higher BLEU-4 score in retrosynthesis, {MoRA}'s dominance in the more chemically stringent EM and FTS metrics underscores its advanced reasoning capabilities beyond superficial textual similarity. We attribute this success to {MoRA}'s core mechanism: by dynamically injecting instance-specific molecular information directly into the frozen LLM's parameter space, this direct parameter-space conditioning allows the LLM to more effectively leverage its vast pre-trained knowledge, yielding more robust and accurate chemical reasoning.

\paragraph{Molecule Description Generation Task.}
In addition, we compare MoRA against a comprehensive set of baselines on two standard molecular description generation benchmarks: Molecular Captioning on {ChEBI-20} and Description Q\&A on {PubChem}. The results are detailed in Table~\ref{tab:description}. Specifically, {MoRA} achieves state-of-the-art results across both tasks. For instance, on the Description Q\&A task, MoRA demonstrates a relative improvement of 86.2\% on the B-4 score over 3D-MoLM. Similarly, for the Molecular Captioning task, our model obtains a 15.9\% relative increase in the B-4 score compared to HIGHT.

\begin{table}[h!]
\centering
\begin{tabular}{@{}lccccc@{}}
\toprule
\textbf{Model}  & \textbf{B-4} $\uparrow$  & \textbf{R-1} $\uparrow$ & \textbf{R-2} $\uparrow$ & \textbf{R-L} $\uparrow$ & \textbf{M} $\uparrow$ \\ 
\midrule
\rowcolor{gray!20}\multicolumn{6}{l}{\textit{Description Q\&A Task}} \\
\midrule
LLaMA2$^1$  & 0.232 & 0.351 & 0.221 & 0.304 & 0.469 \\
3D-MoLM(S)$^1$  & 0.261 & 0.401 & 0.256 & 0.346 & 0.522 \\ 
3D-MoLM(G)$^1$  & 0.261 & 0.401 & 0.259 & 0.350 & 0.519 \\
InstructMol$^1$ & 0.117 & 0.273 & 0.118 & 0.178 & 0.213 \\
UniMoT$^1$      & 0.238  & 0.375  & 0.237  & 0.336  & 0.348 \\
 \midrule
\textbf{MoRA (Ours)} &\textbf{0.486} &\textbf{0.645} &\textbf{0.498} &\textbf{0.578} &\textbf{0.590} \\ 

\bottomrule

\rowcolor{gray!20}\multicolumn{6}{l}{\textit{Molecular Captioning Task}}\\
\midrule
Mol-Instruct$^2$  & 0.171 & 0.331 & 0.203 & 0.289 & 0.271 \\
HIGHT$^2$         & 0.397 & 0.582 & 0.414 & 0.518 & 0.525 \\
InstructMol$^2$   & 0.371 & 0.566 & 0.394 & 0.502 & 0.509 \\

\midrule
\textbf{MoRA (Ours)}  &\textbf{0.460} &\textbf{0.623} &\textbf{0.469} &\textbf{0.563} &\textbf{0.565} \\
\bottomrule
\end{tabular}
\caption{$^1$Results of molecular description generation task on PubChem~\cite{kim2023pubchem}. $^2$Performance of molecule captioning task on the ChEBI-20~\cite{text2mol}. B: BLEU, R: ROUGE, M: METEOR.} 
\label{tab:description}
\end{table}

\begin{table}[h!]
\centering

\begin{tabular}{@{}lcccc@{}}
\toprule
\textbf{Model}  & \textbf{HOMO} $\downarrow$ & \textbf{LUMO} $\downarrow$ & \textbf{GAP} $\downarrow$ & \textbf{Avg.} $\downarrow$ \\ 
\midrule
Alpaca$^1$     & -       & -       & -       & 322.1   \\
LLaMA2$^2$     & 0.737   & 0.864   & 0.515   & 0.751   \\
Vicuna$^2$     & 0.714   & 3.681   & 1.541   & 1.978   \\
Mol-Instruct & 0.0210  & 0.0210  & 0.0203  & 0.0210  \\
HIGHT        & 0.0056  & 0.0065  & 0.0077  & 0.0066  \\
InstructMol  & 0.0048  & {0.0050}  & 0.0061  & {0.0050}  \\
UniMoT       & 0.0042  & 0.0047   & 0.0055 &0.0049 \\
\midrule
\textbf{MoRA (Ours)}  &\textbf{0.0032} &\textbf{0.0037} &\textbf{0.0045} &\textbf{0.0038} \\
\bottomrule
\end{tabular}
\caption{Mean Absolute Error (MAE) on the QM9~\cite{ramakrishnan2014quantum} property prediction task (lower is better). }
\label{tab:qm_results}
\end{table}

\paragraph{Property Prediction Task.}
We further assess the quantitative reasoning capabilities of MoRA on the QM9 quantum property prediction task. As presented in Table~\ref{tab:qm_results}, the results reveal that in-context learning baselines, such as LLaMA2 and Vicuna, struggle significantly, exhibiting substantially higher MAEs. In contrast, InstructMol and UniMoT achieve strong performance. MoRA outperforms all other models across every individual property (HOMO, LUMO, GAP) and in the average MAE. Notably, MoRA achieves an average MAE of 0.0038, marking a 22\% relative improvement over the best-performing baseline, UniMoT (0.0049). This sets a new performance benchmark, underscoring MoRA’s superior ability in precise, molecule-grounded numerical prediction.

\begin{figure*}[t]
    \centering
    \begin{minipage}[t]{0.48\textwidth}
        \centering
        \includegraphics[width=\textwidth]{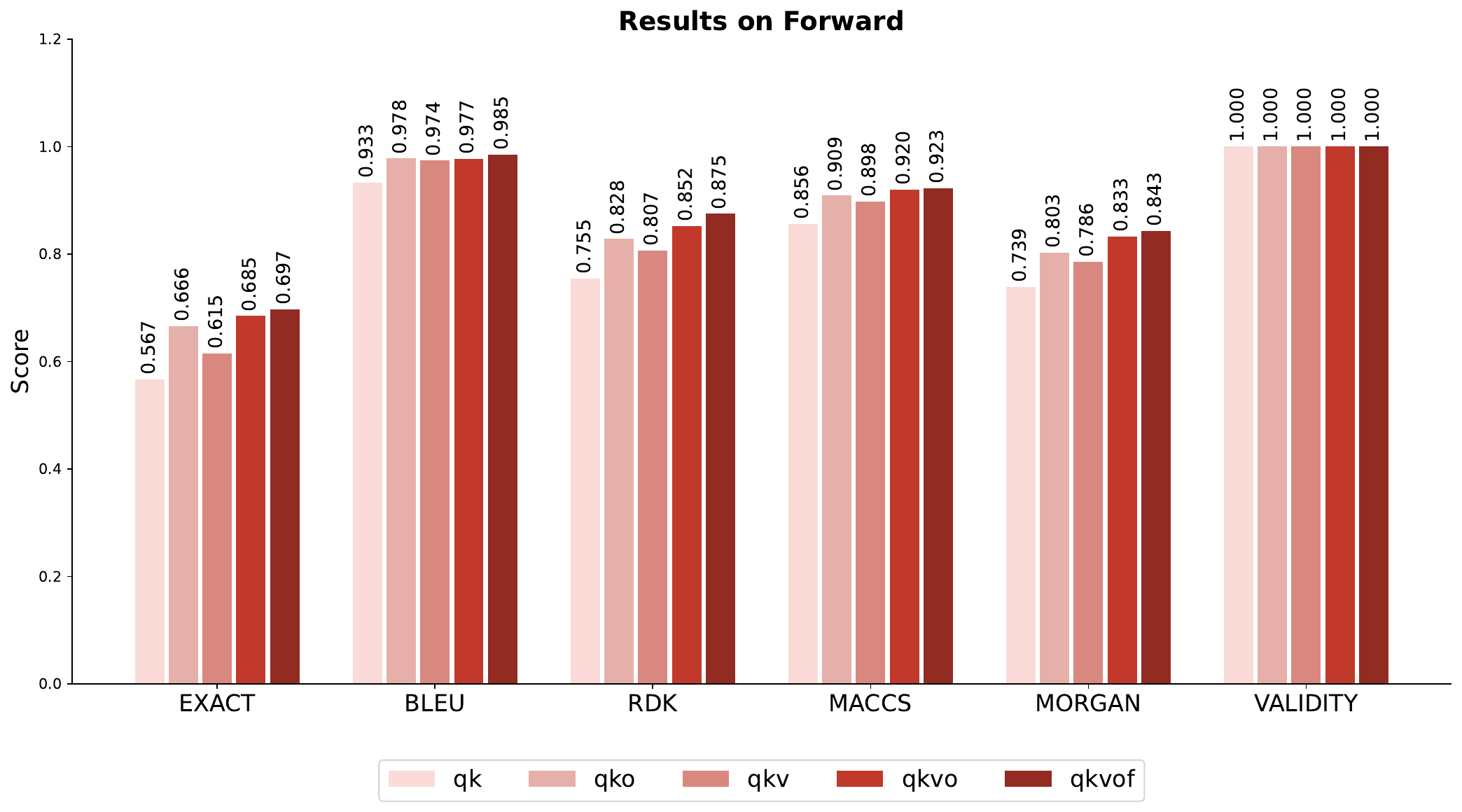}
    \end{minipage}
    \hfill
    \begin{minipage}[t]{0.48\textwidth}
        \centering
        \includegraphics[width=\textwidth]{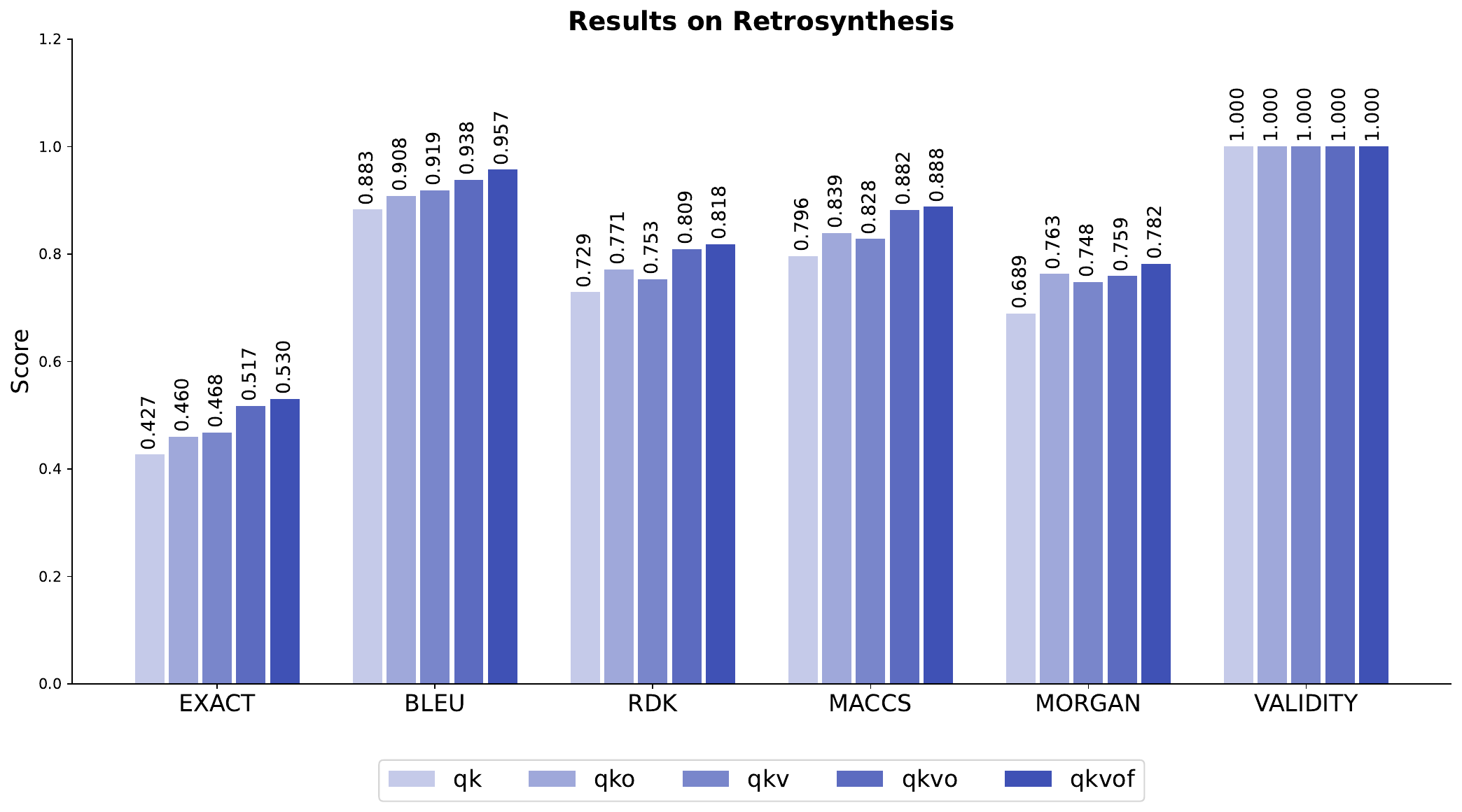}
    \end{minipage}
    \caption{Ablation study on the injection targets for the molecule-aware weights. Targets include the self-attention projections—query (q), key (k), value (v), output (o)—and the feed-forward network (f).}
\label{fig:ablation_weights}
\end{figure*}

\begin{table*}[t]
\centering
\begin{tabular}{l cc cc}
\toprule
\textbf{} & \multicolumn{2}{c}{\textbf{Molecular Tasks (Specialized)}} & \multicolumn{2}{c}{\textbf{General Tasks (Reasoning)}} \\
\cmidrule(lr){2-3} \cmidrule(lr){4-5}
\textbf{Model Variant} & \textbf{Forward.} & \textbf{Retrosynthesis} & \textbf{GSM8K} & \textbf{MMLU} \\
& (Exact $\uparrow$) & (Exact $\uparrow$) & (Accuracy $\uparrow$) & (Accuracy $\uparrow$) \\
\midrule
Vicuna-7B (Inference only)      & - & - & {0.139} & {0.459} \\
MoRA (Task-oriented)   & 0.487 & 0.407 & 0.131 & 0.449 \\ 
{MoRA (Instance-specific)} & {0.697} & {0.530} & 0.137 & 0.455 \\
\bottomrule
\end{tabular}%
\caption{Comprehensive performance evaluation of MoRA as a unified framework. The table contrasts three model variants across two distinct domains: specialized molecular tasks and general reasoning tasks.}
\label{tab:variantscomparison}
\end{table*}

\section{Ablation}

\paragraph{The Impact of Dynamic Instance-specific Adaptation.}
To demonstrate the limitations of static adaptation, we first build a static task-oriented MoRA baseline, where a single fixed low-rank adapter is trained on the molecular dataset and applied consistently to all instances. As shown in Table \ref{tab:variantscomparison}, the Original Vicuna model is unable to perform the specialized molecular tasks. While the static task-oriented variant successfully acquires these skills, our dynamic instance-specific MoRA proves to be significantly more effective. Specifically, the instance-specific variant achieves much higher exact match scores of 0.697 in forward synthesis and 0.530 in retrosynthesis. Crucially, this high performance on specialized tasks is achieved while maintaining general reasoning abilities comparable to the original model, underscoring the superiority of dynamic, instance-specific conditioning over static adaptation.

\begin{table}[]
\centering
\begin{tabular}{l cc cc}
\toprule
 & \multicolumn{2}{c}{\textbf{Forward.}} & \multicolumn{2}{c}{\textbf{Retrosyn.}} \\
\cmidrule(lr){2-3} \cmidrule(lr){4-5}
\textbf{Blocks} 
& Exact $\uparrow$ & BLEU $\uparrow$& Exact $\uparrow$  & BLEU $\uparrow$ \\
\midrule
N = 2  & 0.663 & 0.967 & 0.493  & 0.928 \\
N = 4  & 0.645 & 0.969 & 0.508  & 0.944 \\
N = 8  & 0.697 & 0.985 & 0.533  & 0.957 \\
N = 16 & 0.689 & 0.980 & 0.530  & 0.955 \\
\bottomrule
\end{tabular}%

\caption{Performance metrics for models with a varying number of blocks (N). Arrows indicate if higher ($\uparrow$) or lower ($\downarrow$) values are better. Forward. means Forward Reaction Prediction task, and Retrosyn. means Retrosynthesis task.}
\label{tab:block_performance_alt}
\end{table}

\paragraph{The type of weights.}
We conduct an ablation study to identify the optimal injection targets for MoRA's dynamic weights within a standard LLM decoder block. The candidate targets include the multi-head attention matrices — query ({q}), key ({k}), value ({v}), output projection ({o}), and the feed-forward network ({f}). As shown in Figure~\ref{fig:ablation_weights}, the optimal performance is achieved by the \texttt{qkvof} configuration, which concurrently injects weights into all of these components. This finding suggests that holistic modulation of the entire computational block, encompassing both the self-attention and feed-forward layers, is essential for achieving maximal efficacy in structural conditioning.

\paragraph{The number of blocks of the perceptual weights generator.}
We investigate the effect of the depth of the Molecule-Aware Weight Generator (MAW-Gen) controlled by the number of its transformer blocks (N). 
As shown in Table~\ref{tab:block_performance_alt}, performance generally trends upward as \texttt{N} increases from 2 to 8, with all metrics reaching their peak at \texttt{N=8}.  A shallower model (\texttt{N=2} or \texttt{N=4}) appears to lack sufficient expressive capacity, resulting in lower scores.  Therefore, we adopt \texttt{N=8} as it consistently yields the best results across both tasks.

\section{Conclusion}
In this paper, we introduced MoRA, a novel multi-modal framework that shifts from static, task-oriented fine-tuning to dynamic, instance-specific adaptation. Instead of permanently altering the LLM, MoRA employs a Molecule-Aware Weight Generator to produce unique low-rank adaptation weights for each input molecule on-the-fly. These weights are injected into the frozen LLM, enabling structure-aware computation while preserving the model’s core knowledge. Our experiments demonstrate that MoRA achieves state-of-the-art performance on key molecular tasks. MoRA effectively mitigates the conflict between domain specialization and general intelligence, offering a new paradigm for molecular multi-modal assistants. In future work, we will extend our experiments to other molecular modalities.

\section{Appendix}

\subsection{A. Detailed Experimental Settings}

\paragraph{A1. Details of Benchmarks for Evaluation}

\paragraph{Molecular Description Generation.}
This task evaluates a model's ability to generate informative text descriptions for a given molecule. The descriptions should cover key aspects such as chemical properties, functional groups, and biological roles. We use two main datasets for this task. (1) ChEBI-20 provides rich, detailed descriptions for molecules, focusing on biochemical entities. (2) The PubChem Description Q\&A dataset is structured in a question-and-answer format, testing the model's ability to retrieve specific factual information about a molecule.
\paragraph{Forward Reaction Prediction.}
In this task, the model predicts the likely product of a chemical reaction given the reactants and reagents. The model receives input as SMILES strings representing the starting materials and must generate the SMILES string for the resulting product. Our evaluation is based on the widely used USPTO dataset, which contains a large collection of chemical reactions from U.S. patents.
\paragraph{Retrosynthesis.}
Retrosynthesis is the reverse of forward reaction prediction. Given a target product molecule, the model's goal is to identify and generate the SMILES strings of the necessary reactants for its synthesis. This task challenges the model's understanding of chemical reaction pathways. We use the USPTO 500K dataset, a large-scale benchmark specifically curated for single-step retrosynthesis.
\paragraph{Reagent Prediction.}
This task assesses the model's ability to identify the correct reagents needed to facilitate a specific chemical transformation. The model is provided with both the reactants and the final product, and it must predict the appropriate reagent(s). For this task, we utilize the USPTO 500 MT dataset.
\paragraph{Property Prediction (Regression).}
This task measures the model's performance in predicting quantitative molecular properties. We focus on quantum mechanical properties, specifically predicting the energy of the Highest Occupied Molecular Orbital (HOMO) and the Lowest Unoccupied Molecular Orbital (LUMO). The model takes a molecule's SMILES representation as input and outputs a continuous numerical value for the target property. We use the QM9 dataset for this evaluation.

\paragraph{A2. Implementation Details}
MoRA is built upon the Vicuna-7B model, whose original parameters are kept frozen during training. The architecture incorporates a frozen GNN-based graph encoder and our proposed trainable module, the Molecule-Aware Weight Generator (MAW-Gen). The graph encoder processes molecular structures, while the MAW-Gen is specifically designed to produce molecule-conditioned, low-rank updates. This generator consists of \(N=8\) decoder blocks and \(k=4\) learnable molecular queries. The generated low-rank matrices, with a rank of \(r=64\), are injected into the query, key, value, and output projection matrices ($\mathbf{W}_q, \mathbf{W}_k, \mathbf{W}_v, \mathbf{W}_o$) within each self-attention layer, as well as the feed-forward network (FFN) projections ($\mathbf{f}$). To ensure training stability, we follow established practices for LoRA-based methods and initialize the linear projection head in MAW-Gen, which produces the update matrix, with zeros. The model is optimized using the AdamW optimizer to minimize the standard cross-entropy loss for text generation. Key hyperparameters and configuration details are summarized in Table \ref{tab:mora_implementation_details}.

\begin{table}[h!]
\centering
\renewcommand{\arraystretch}{1.2}
\begin{tabular}{@{}lc@{}}
\toprule
\textbf{Configuration} & \textbf{MoRA} \\ \midrule
Molecule Encoder & MoleculeSTM \\
Molecule Emb. Dim. & 300\\
LLM Base & Vicuna-7B \\
PEFT Method & MoRA (Ours) \\
{MoRA Injection Targets} & {q, k, v, o, f} \\
{MoRA-Rank ($r$)} & {64} \\
{MoRA-Alpha ($\alpha$)} & {64} \\
{MAW-Gen Blocks ($N$)} & {8} \\
{Molecular Queries ($k$)} & {4} \\
\midrule
\multicolumn{2}{@{}l}{\textbf{Training Hyperparameters}} \\
\midrule
Training Stages & Single Stage Fine-tuning \\
Epochs & 10-30 \\
Optimizer & AdamW \\
Batch Size & 128 (Global) \\
Learning Rate & 2e-5 \\
LR Scheduler & Cosine \\
Warm-up & 3\% of total steps \\
Weight Decay & 0.0 \\
Precision & bfloat16 \\
GPU Usage & 8 NVIDIA A800s \\ \bottomrule
\end{tabular}
\caption{Implementation details for our MoRA model.}
\label{tab:mora_implementation_details}
\end{table}

\begin{table*}[]
  \centering
  \captionsetup{skip=5pt, font=small, labelfont=bf} 

  \label{tab:datasets_corrected}
  
  \renewcommand{\arraystretch}{1.3} 
  
  \begin{tabular}{@{}lllr@{}}
    \toprule
    \textbf{TASKS} & \textbf{Dataset} & \textbf{Source} & \multicolumn{1}{c}{\textbf{Split (Train/Val/Test)}} \\ 
    \midrule
    
    \multirow{2}{*}{Molecule Description Generation} 
    & ChEBI-20 (Molecular Captioning) & ChEBI / PubChem & 26,420 / 3,295 / 3,295 \\
    & PubChem (Description Q\&A) & PubChem & 56,885 / - / 10,000 \\ 
    
    Forward Prediction & USPTO & USPTO & 124,384 / - / 1,000 \\ 
    Retrosynthesis & USPTO 500K & USPTO & 128,684 / - / 1,000 \\
    
    \multirow{1}{*}{Property Prediction (Regression)} 
    & QM9 (HOMO/LUMO) & GDB-17 & 360,113 / - / 1,987 \\
    \bottomrule
  \end{tabular}
    \caption{Summary of Datasets for Molecular Tasks.}
\end{table*}

\begin{table*}[h!]
\centering
\renewcommand{\arraystretch}{1.2} 
\begin{tabular}{@{}l ccccccc@{}}
\toprule
\textbf{Model} & \textbf{Exact ↑} & \textbf{BLEU ↑} & \textbf{Levenshtein ↓} & \textbf{RDK ↑} & \textbf{MACCS ↑} & \textbf{Morgan ↑} & \textbf{Validity ↑} \\ \midrule
Vicuna  & 0.000          & 0.010          & 27.948          & 0.038          & 0.002          & 0.001          & 0.007          \\
LLaMA2   & 0.000          & 0.283          & 53.510          & 0.136          & 0.294          & 0.106          & 1.000          \\
Mol-Instruction  & 0.044          & 0.224          & 23.167          & 0.237          & 0.364          & 0.213          & 1.000          \\
HIGHT    & 0.067          & 0.482          & 27.167          & 0.462          & 0.346          & 0.303          & 1.000          \\
InstructMol   & 0.129          & 0.610          & 19.664          & 0.444          & 0.539          & 0.400          & 1.000          \\
\midrule
\textbf{MoRA}          & 0.158          & 0.664          & 17.633          & 0.450          & 0.549          & 0.418          & 1.000          \\ \bottomrule
\end{tabular}
\caption{Performance comparison on the Reagent Prediction Task. Our model (MoRA) is compared against various baselines. All scores for MoRA are highlighted. Arrows (↑/↓) indicate whether higher or lower scores are better.}
\label{tab:reagent_prediction}
\end{table*}

\paragraph{A3. Details of Evaluation Metrics}
We use a range of established metrics to evaluate our model's performance across different tasks. The direction of the arrow (↑/↓) indicates whether a higher or lower score is better.

\paragraph{Chemical Reaction Prediction.}
For tasks involving SMILES generation (Forward Prediction, Retrosynthesis, Reagent Prediction), we use the following metrics:
\begin{itemize}

    \item \textbf{Exact Match (↑):} The percentage of predictions where the canonicalized SMILES string of the generated molecule is identical to the ground truth. This is a strict measure of accuracy.
    \item \textbf{BLEU (↑):} {Measures the n-gram overlap between the generated and reference SMILES strings. Although typically used for natural language, it serves as a proxy for textual similarity in SMILES generation.}
    \item \textbf{Levenshtein Score (↓):} Measures the minimum number of single-character edits (insertions, deletions, substitutions) needed to transform the predicted SMILES into the ground-truth SMILES. It quantifies the textual difference between the strings.
    \item \textbf{Structural Similarity (↑):} We assess the structural likeness between predicted and true molecules using Tanimoto similarity calculated from different molecular fingerprints: \textbf{MACCS}, \textbf{RDKit}, and \textbf{Morgan} fingerprints. A higher score indicates greater structural resemblance.
    \item \textbf{Validity (↑):} The percentage of generated SMILES strings that represent chemically valid and parsable molecules. This is a fundamental check for the model's chemical correctness.
\end{itemize}

\paragraph{Molecule Description Generation.}
To assess the quality of the generated text, we employ a comprehensive set of standard metrics that compare the predicted descriptions against reference texts:
\begin{itemize}
    \item \textbf{BLEU-4 (B-4) (↑):} Measures the precision of n-grams (up to 4-grams), evaluating the fluency and adequacy of the generated text by checking for co-occurring word sequences.
    \item \textbf{ROUGE (R) (↑):} We report three variants to measure recall. \textbf{ROUGE-1 (R-1)} and \textbf{ROUGE-2 (R-2)} compute the overlap of unigrams and bigrams, respectively. \textbf{ROUGE-L (R-L)} is based on the Longest Common Subsequence (LCS) and captures sentence-level structural similarity.
    \item \textbf{METEOR (M) (↑):} A more advanced metric that creates an alignment between the generated and reference texts, considering not only exact matches but also stems and synonyms for a more semantically-aware evaluation.
\end{itemize}

\paragraph{Property Prediction (Regression).}
For the quantitative prediction of quantum properties, we evaluate the model's accuracy on HOMO energy, LUMO energy, and the HOMO-LUMO Gap.
\begin{itemize}
    \item \textbf{Mean Absolute Error (MAE) (↓):} We report the MAE for each property (\textbf{HOMO}, \textbf{LUMO}, \textbf{GAP}) to measure the specific prediction error for each task. We also compute the \textbf{Average MAE (Avg. MAE)}, which is the mean of these three MAE values, to provide a single, overall performance score.
    \begin{equation}
        \text{MAE} = \frac{1}{n} \sum_{i=1}^{n} |y_i - \hat{y}_i|
    \end{equation}
    
\end{itemize}

\subsection{B. Additional Experimental Results}
\paragraph{B1. Reagent Prediction.}
We present the performance of our MoRA model on the Reagent Prediction task, with detailed results provided in Table \ref{tab:reagent_prediction}. The findings indicate that MoRA achieves highly competitive performance compared to established baseline models. Specifically, our model significantly surpasses foundational models like Vicuna and LLaMA2 across all metrics. Furthermore, MoRA consistently outperforms more specialized models such as Mol-Instruction and InstructMol in key areas including Exact Match, BLEU score, and Levenshtein distance, while ensuring 100\% validity of the generated molecules. These results underscore the effectiveness of our approach in accurately predicting chemical reagents from reaction context.

\paragraph{B2. Analysis of the Computational Cost}


\begin{figure*}[t]
    \centering
    \begin{subfigure}[b]{0.48\textwidth}
        \includegraphics[width=\textwidth]{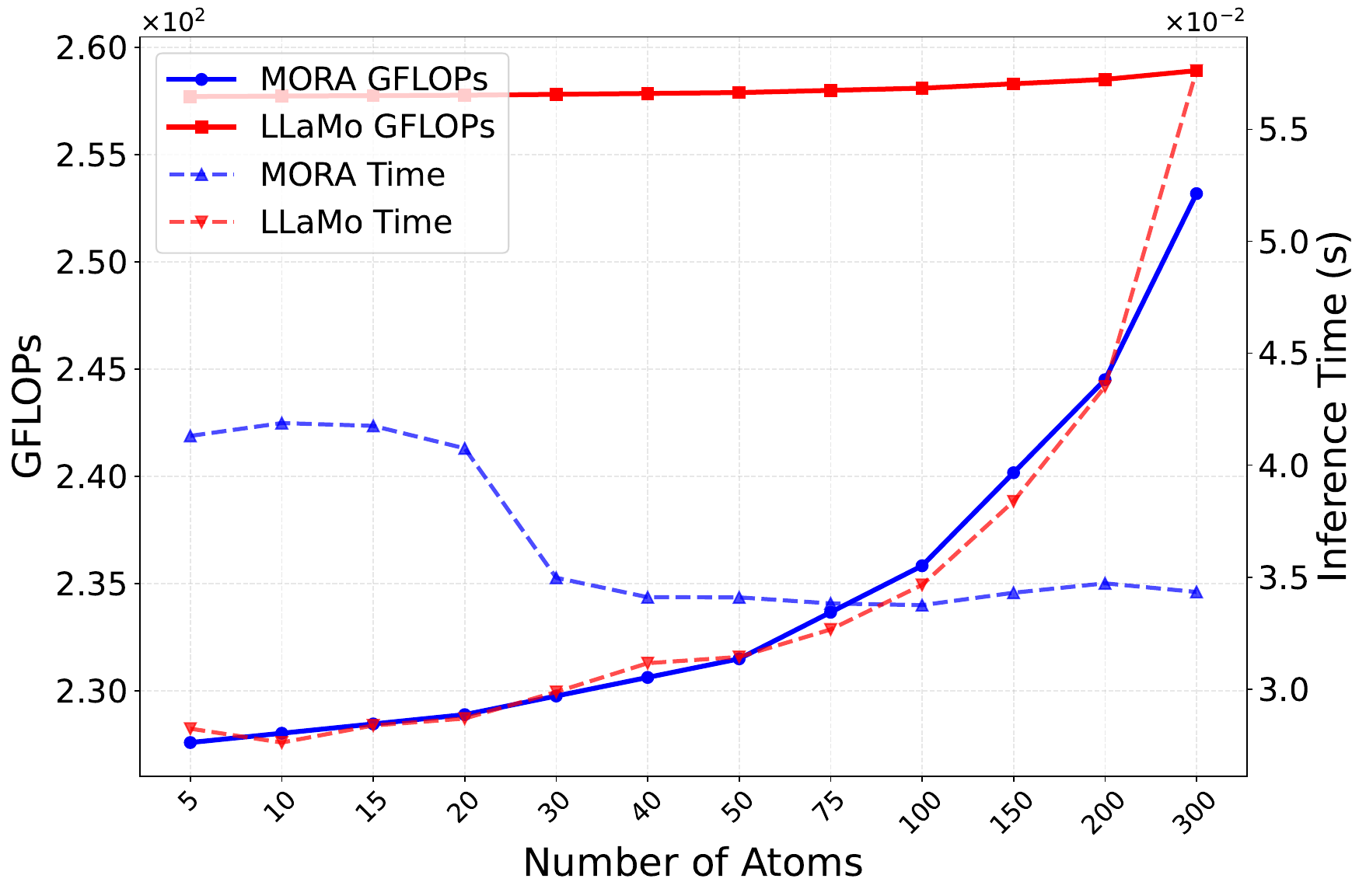} 
        \caption{GFLOPS and Inference Time vs. Number of Atoms.}
        \label{fig:atoms-gflops-time}
    \end{subfigure}
    \hfill
    \begin{subfigure}[b]{0.48\textwidth}
        \includegraphics[width=\textwidth]{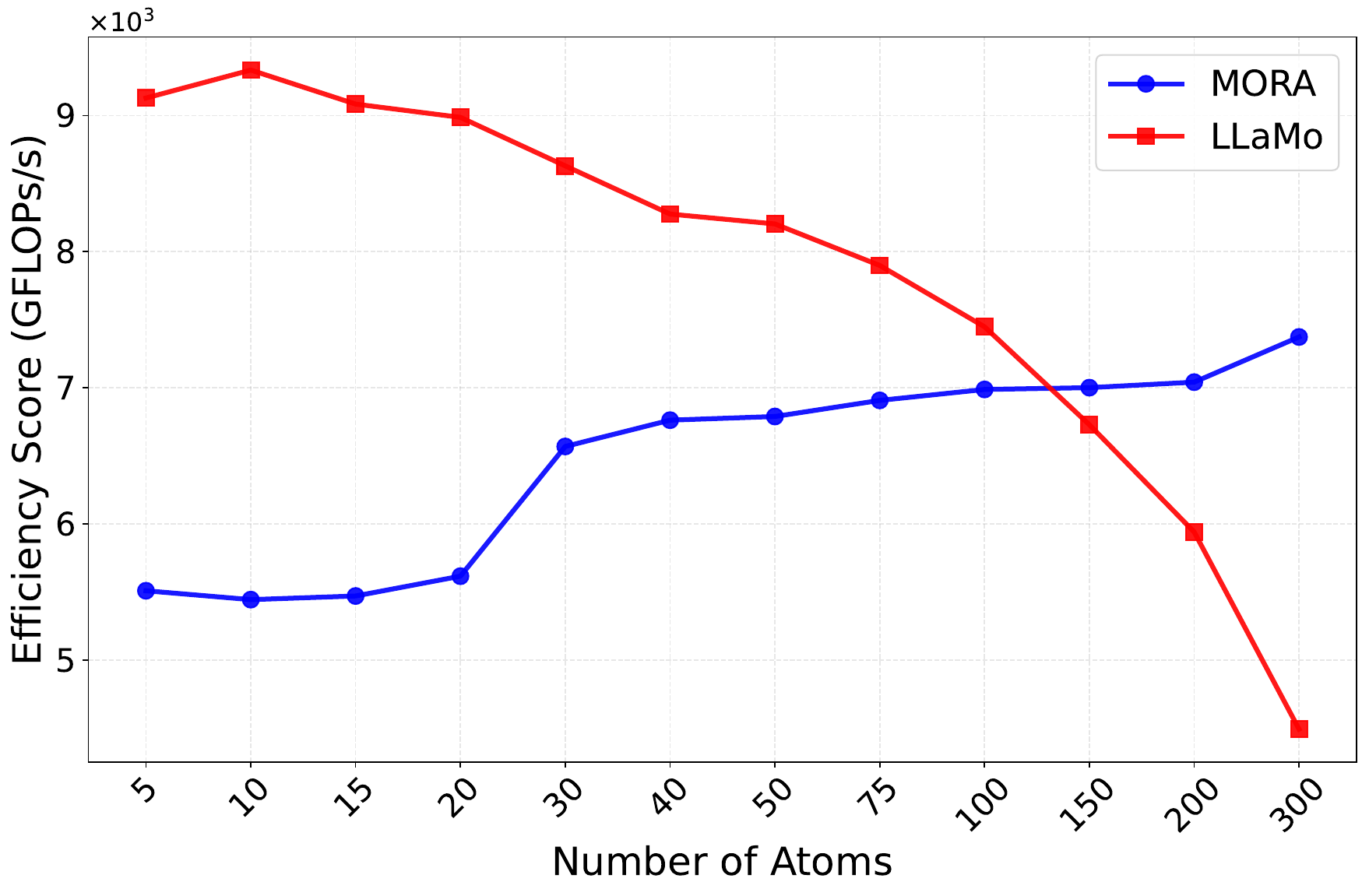} 
        \caption{Efficiency Score (GFLOPS/s) vs. Number of Atoms.}
        \label{fig:atoms-efficiency}
    \end{subfigure}
    \caption{Computational efficiency analysis as a function of {molecular complexity}. (a) Total GFLOPS and inference time scaling with the number of atoms. (b) The corresponding efficiency score, highlighting resource utilization.}
    \label{fig:efficiency-vs-atoms}
\end{figure*}

\begin{figure*}[t]
    \centering
    \begin{subfigure}[b]{0.48\textwidth}
        \includegraphics[width=\textwidth]{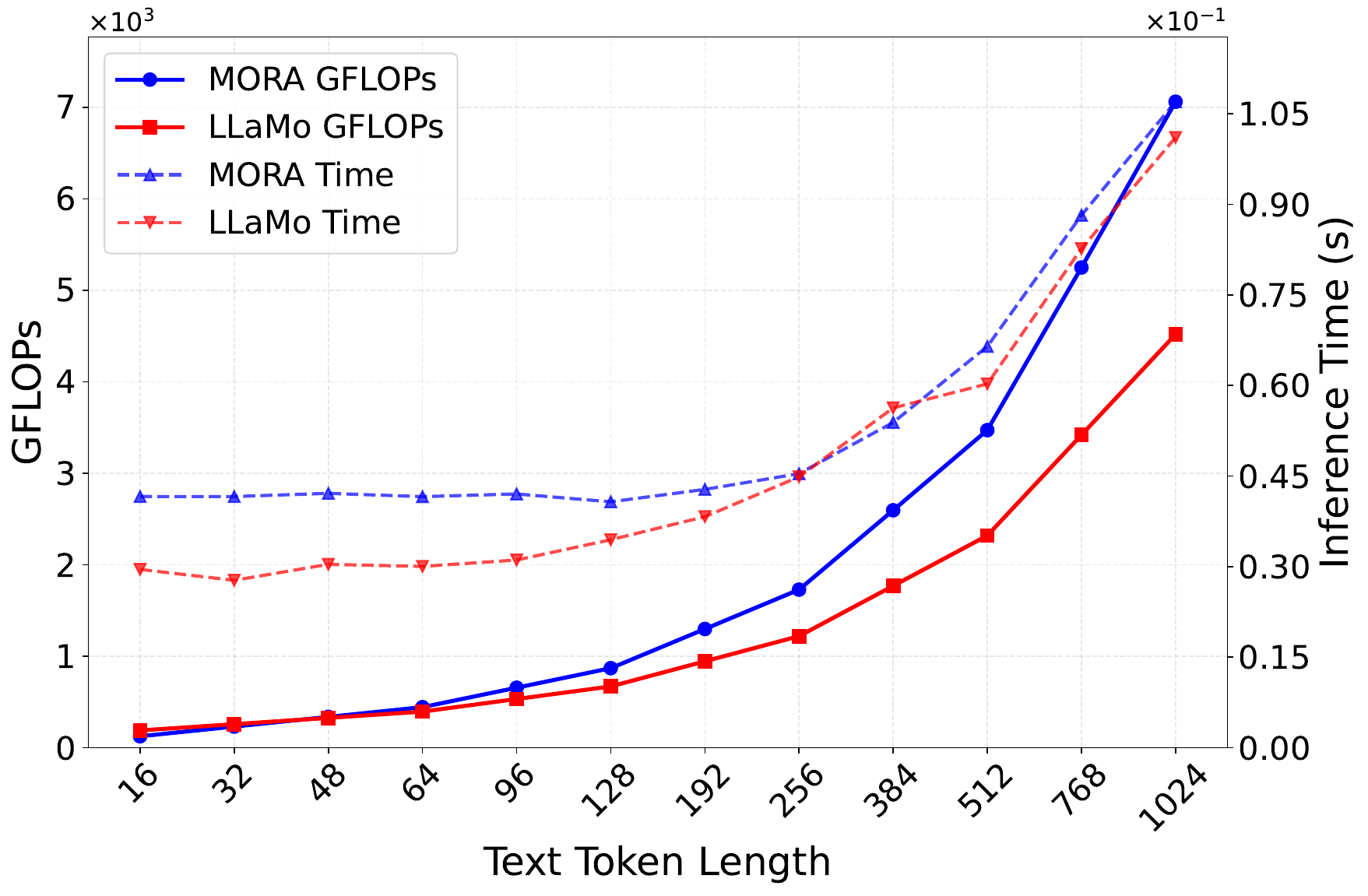} 
        \caption{GFLOPS and Inference Time vs. Text Token Length.}
        \label{fig:text-gflops-time}
    \end{subfigure}
    \hfill
    \begin{subfigure}[b]{0.48\textwidth}
        \includegraphics[width=\textwidth]{LaTeX/pic/appendix/efficiency_vs_atoms.pdf} 
        \caption{Efficiency Score (GFLOPS/s) vs. Text Token Length.}
        \label{fig:text-efficiency}
    \end{subfigure}
    \caption{Computational efficiency analysis as a function of {text input length}. (a) Total GFLOPS and inference time scaling with token length. (b) The corresponding efficiency score, reflecting throughput on longer sequences.}
    \label{fig:efficiency-vs-text}
\end{figure*}

To evaluate the computational profile of our framework, we conduct a detailed efficiency and scalability analysis, comparing MoRA against LLaMo, a representative model performing modality alignment in the input space. The results, presented in Figure~\ref{fig:efficiency-vs-atoms} and Figure~\ref{fig:efficiency-vs-text}, reveal the distinct computational trade-offs of parameter-space modulation versus input-space alignment.

\paragraph{Scalability with Molecular Complexity.}
We first assess how performance scales with the size of the input molecule, a critical factor for real-world chemical applications. As shown in Figure~\ref{fig:efficiency-vs-atoms}(a), the two frameworks exhibit fundamentally different scaling behaviors. LLaMo's inference time remains largely constant regardless of the number of atoms. This is because its computational cost is dominated by processing a fixed-length tokenized representation of the molecule. In contrast, MoRA's inference time increases linearly with molecular size, reflecting the computational cost of its GNN encoder and the Molecule-Aware Weight Generator (MAW-Gen), which directly operate on the graph structure. This architectural difference leads to important implications for efficiency, as illustrated in Figure~\ref{fig:efficiency-vs-atoms}(b). LLaMo's efficiency, measured as GFLOPS per second, is high for small molecules but degrades sharply as complexity increases. This indicates that its substantial computational overhead (the entire LLM) becomes increasingly underutilized for the core task of interpreting larger molecular structures. Conversely, MoRA's efficiency, while starting lower due to the initial overhead of its generator, remains stable and robust across all molecule sizes. 

\paragraph{Scalability with Text Length.}
Next, we examine scalability with respect to the length of the textual input, which is crucial for tasks involving long instructions or detailed generated answers. Figure~\ref{fig:efficiency-vs-text}(a) shows that for both models, GFLOPS and inference time increase with sequence length, an expected characteristic of autoregressive transformers. However, their efficiency profiles, shown in Figure~\ref{fig:efficiency-vs-text}(b), reveal a key advantage of our approach. Both models become more efficient on longer sequences, likely due to improved hardware parallelization. Notably, MoRA's efficiency improves at a significantly steeper rate, surpassing LLaMo for sequences longer than approximately 96 tokens. This suggests that MoRA's architecture is better optimized for sustained, long-form text processing, making it more effective for complex dialogue or descriptive tasks.

In summary, our analyses reveal a clear and compelling computational trade-off. LLaMo is faster for very simple tasks involving small molecules. However, this advantage quickly vanishes as input complexity grows, where its architecture proves inefficient. MoRA, on the other hand, demonstrates superior scalability. While it carries a slightly higher baseline cost from its specialized components, its performance scales gracefully with both molecular complexity and text length. 

\subsection{C. Pseudocode of MoRA}
We present the pseudocode for MoRA in Algorithm \ref{alg:mora}. 
The MoRA framework dynamically adapts a frozen Large Language Model (LLM) for each input molecule $\mathcal{G}$. First, a frozen Graph Neural Network (GNN) encodes $\mathcal{G}$ into node embeddings. A trainable Molecule-Aware Weight Generator (MAW-Gen) then distills this structural information into a fixed set of molecular queries using cross-attention. These queries are projected to generate a molecule-specific, low-rank weight update $\mathbf{W}_{\text{mol}} = \Delta\mathbf{A} \cdot \mathbf{W}_{\text{proj}}$. This update additionally modulates the LLM's original attention weights ($\hat{\mathbf{W}} \leftarrow \mathbf{W} + \mathbf{W}_{\text{mol}}$), tailoring the model for the given molecule before it processes the text input. The MAW-Gen parameters are optimized end-to-end to minimize the language modeling loss. 

\subsection{D. Open Data and Source Code}
Our code and experimental datasets are detailed in the supplementary materials. Once the paper is accepted, we will open-source it on GitHub.

\subsection{E. Case Study}
Table~\ref{tab:desc_comparison} compares the ground truth descriptions with those generated by MoRA.
\subsection{F. Limitations and Future Work }
While MoRA demonstrates significant advantages in molecular representation learning, we identify several key areas for future research that address its current limitations. First, a computational trade-off exists: while MoRA's expressive per-molecule weight generator enhances fidelity over static input-space projectors, it incurs significant overhead. Optimizing this generator via methods like model compression or architectural search is critical for broadening its applicability in time-sensitive scenarios. Second, the current 2D graph representation limits the model's ability to capture fine-grained chemical details. This is particularly problematic for stereoisomers, as molecules with identical 2D connectivity (e.g., enantiomers) can have profoundly different biological activities, a crucial distinction that a 2D-based model fundamentally cannot make. 

\begin{algorithm}[]
\caption{MoRA: Molecule-aware Low-Rank Adaptation Framework}
\label{alg:mora}
\KwIn{Training dataset $\mathcal{D} = \{(\mathcal{G}_i, I_i, A_i)\}_{i=1}^N$, Frozen GNN encoder $g(\cdot)$, Frozen LLM with parameters $\theta_{LLM}$}
\KwOut{Trained MAW-Gen parameters $\psi$}
\BlankLine
\textbf{Initialize:}\\
\Indp
Molecular queries $\mathbf{Q}_{\text{learn}} \in \mathbb{R}^{k \times d_{\text{model}}}$\\
Transformer decoder with $N$ layers\\
Projection layer $\mathbf{W}_{\text{FC}} \in \mathbb{R}^{d_{\text{model}} \times (d_{\text{llm}} \cdot r)}$\\
Shared projection $\mathbf{W}_{\text{proj}} \in \mathbb{R}^{r \times d_{\text{llm}}}$\\
Components to modulate $\mathcal{C} = \{W_q, W_k, W_v, W_o\}$ across $L_{\text{adapt}}$ layers
\Indm
\BlankLine
\For{epoch $= 1$ \KwTo $E$}{
    \For{batch $(\mathcal{G}, I, A)$ \textbf{in} $\mathcal{D}$}{
        \tcp{Step 1: Encode molecular graph}
        $\mathbf{H}_\mathcal{G} \leftarrow g(\mathcal{G})$ \tcp*{Node embeddings $\{\mathbf{h}_v^{(L)}\}_{v \in \mathcal{V}}$}
        \BlankLine
        \tcp{Step 2: Cross-attention distillation}
        $\mathbf{Q} \leftarrow \mathbf{Q}_{\text{learn}}$\\
        \For{$l = 1$ \KwTo $N$}{
            $\mathbf{Q} \leftarrow \text{SelfAttention}(\mathbf{Q})$\\
            $\mathbf{Q} \leftarrow \text{CrossAttention}(\mathbf{Q}, \mathbf{H}_\mathcal{G})$\\
            $\mathbf{Q} \leftarrow \text{FFN}(\mathbf{Q})$
        }
        $\mathbf{Q}_{\text{out}} \leftarrow \mathbf{Q}$ \tcp*{Distilled queries}
        \BlankLine
        \tcp{Step 3: Generate low-rank weight updates}
        \For{layer $i = 1$ \KwTo $L_{\text{adapt}}$}{
            \For{component $c \in \mathcal{C}$}{
                $\mathbf{q}_{i,c} \leftarrow \mathbf{Q}_{\text{out}}[i \cdot |\mathcal{C}| + c]$ \tcp*{Select query}
                $\Delta\mathbf{A}_{i,c} \leftarrow \text{Reshape}(\mathbf{q}_{i,c} \cdot \mathbf{W}_{\text{FC}})$ \tcp*{Shape: $d_{\text{llm}} \times r$}
                $\mathbf{W}_{\text{mol},i,c} \leftarrow \Delta\mathbf{A}_{i,c} \cdot \mathbf{W}_{\text{proj}}$ \tcp*{Shape: $d_{\text{llm}} \times d_{\text{llm}}$}
            }
        }
        \BlankLine
        \tcp{Step 4: Inject weights into frozen LLM}
        \For{layer $i = 1$ \KwTo $L_{\text{adapt}}$}{
            \For{component $c \in \mathcal{C}$}{
                $\hat{\mathbf{W}}_{i,c} \leftarrow \mathbf{W}_{i,c} + \mathbf{W}_{\text{mol},i,c}$ \tcp*{Parameter modulation}
            }
        }
        \BlankLine
        \tcp{Step 5: Forward pass with adapted LLM}
        $\hat{A} \leftarrow \text{LLM}_{\text{adapted}}(I; \{\hat{\mathbf{W}}_{i,c}\})$ \tcp*{Only text input}
        \BlankLine
        \tcp{Step 6: Compute loss and update MAW-Gen}
        $\mathcal{L} \leftarrow -\sum_{t=1}^{|A|} \log P(a_t | a_{<t}, I, \mathcal{G}; \psi)$\\
        Update $\psi = \{\mathbf{Q}_{\text{learn}}, \mathbf{W}_{\text{FC}}, \mathbf{W}_{\text{proj}}$ using $\nabla \mathcal{L}$
    }
}
\Return{Trained MAW-Gen with parameters $\psi$}
\end{algorithm}

\begin{table*}[]
\centering
\begin{tabular}{>{\centering\arraybackslash}m{0.25\textwidth} m{0.35\textwidth} m{0.35\textwidth}}
\toprule
\textbf{Molecule} & \textbf{Ground Truth} & \textbf{MoRA} \\
\midrule
\includegraphics[width=\linewidth, height=3cm, keepaspectratio]{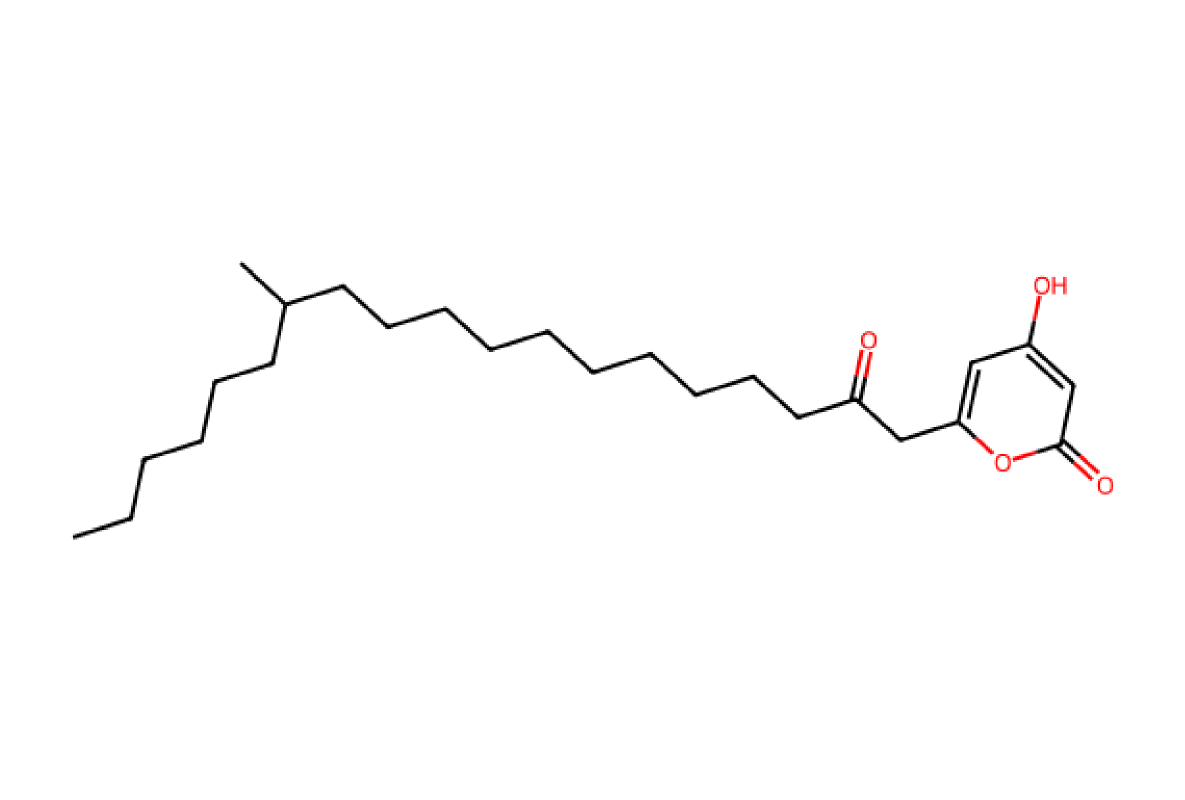} & 
The molecule is an \textcolor{matchcolor}{\textbf{acyl tetraketide pyran-2-one}} that is \textcolor{matchcolor}{\textbf{4-hydroxy-2H-pyran-2-one}} in which the hydrogen at \textcolor{matchcolor}{\textbf{position 6}} is replaced by a \textcolor{matchcolor}{\textbf{13-methyl-2-oxo}}nopadecyl group. & 
The molecule is an \textcolor{matchcolor}{\textbf{acyl tetraketide pyran-2-one}} that is \textcolor{matchcolor}{\textbf{4-hydroxy-2H-pyran-2-one}} in which the hydrogen at \textcolor{matchcolor}{\textbf{position 6}} is replaced by a \textcolor{matchcolor}{\textbf{13-methyl-2-oxo}}\textcolor{diffcolor}{tetradecyl} group. \\
\midrule
\includegraphics[width=\linewidth, height=3cm, keepaspectratio]{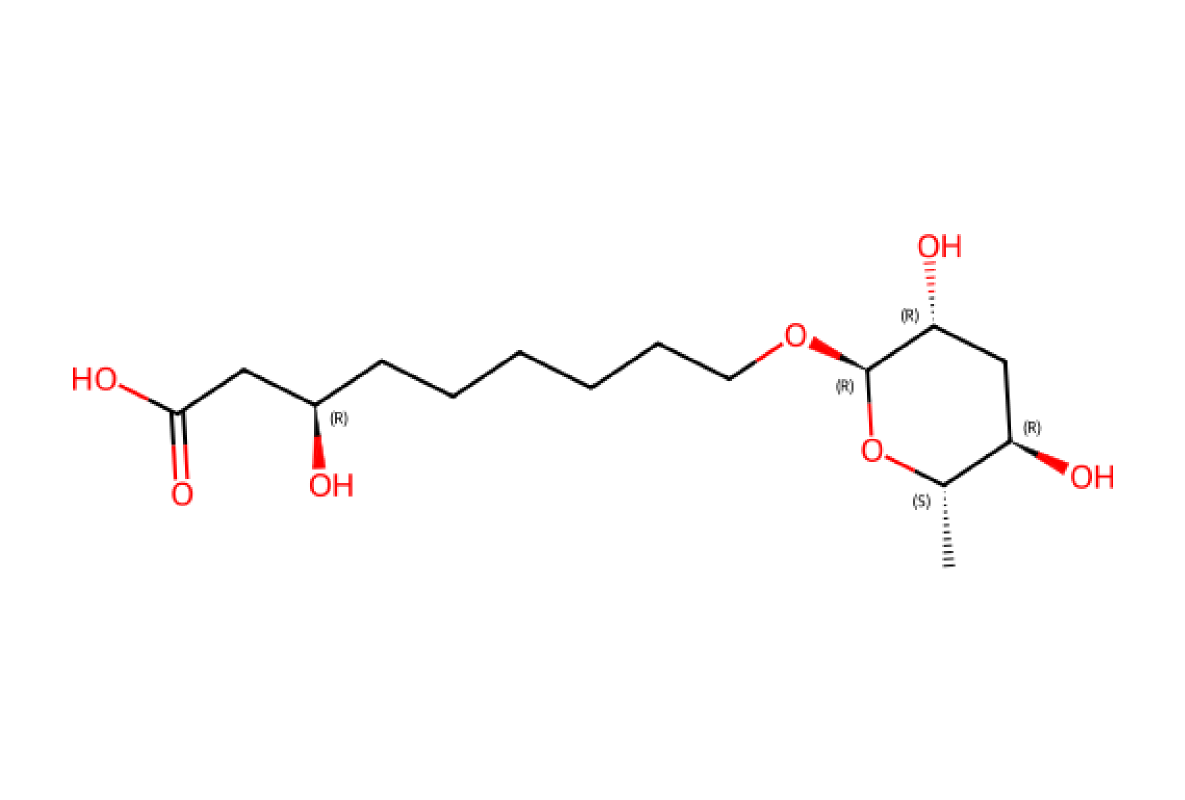} & 
The molecule is an \textcolor{matchcolor}{\textbf{omega-hydroxy fatty acid ascaroside}} that is oscr\#10 in which the \textcolor{matchcolor}{\textbf{pro-R hydrogen beta to the carboxy group}} is replaced by a \textcolor{matchcolor}{\textbf{hydroxy group}}. It is a metabolite of the nematode Caenorhabditis elegans. It has a role as a Caenorhabditis elegans metabolite. It is an \textcolor{matchcolor}{\textbf{omega-hydroxy fatty acid ascaroside}}, a \textcolor{matchcolor}{\textbf{3-hydroxy carboxylic acid}} and a \textcolor{matchcolor}{\textbf{monocarboxylic acid}}. It derives from an oscr\#10 and a \textcolor{matchcolor}{\textbf{(3R)-3}},9-dihydroxynonanoic acid. It is a conjugate acid of a bhos\#10(1-).
&
The molecule is an \textcolor{matchcolor}{\textbf{omega-hydroxy fatty acid ascaroside}} that is \textcolor{diffcolor}{oscr\#18} in which the \textcolor{matchcolor}{\textbf{pro-R hydrogen beta to the carboxy group}} is replaced by a \textcolor{matchcolor}{\textbf{hydroxy group}}. It is a metabolite of the nematode Caenorhabditis elegans. It has a role as a Caenorhabditis elegans metabolite. It is an \textcolor{matchcolor}{\textbf{omega-hydroxy fatty acid ascaroside}}, a \textcolor{matchcolor}{\textbf{3-hydroxy carboxylic acid}} and a \textcolor{matchcolor}{\textbf{monocarboxylic acid}}. It derives from an \textcolor{diffcolor}{oscr\#18} and a \textcolor{matchcolor}{\textbf{(3R)-3}},\textcolor{diffcolor}{11-dihydroxyundecanoic} acid. It is a conjugate acid of a \textcolor{diffcolor}{bhos\#18(1-)}.\\
\midrule
\includegraphics[width=\linewidth, height=3cm, keepaspectratio]{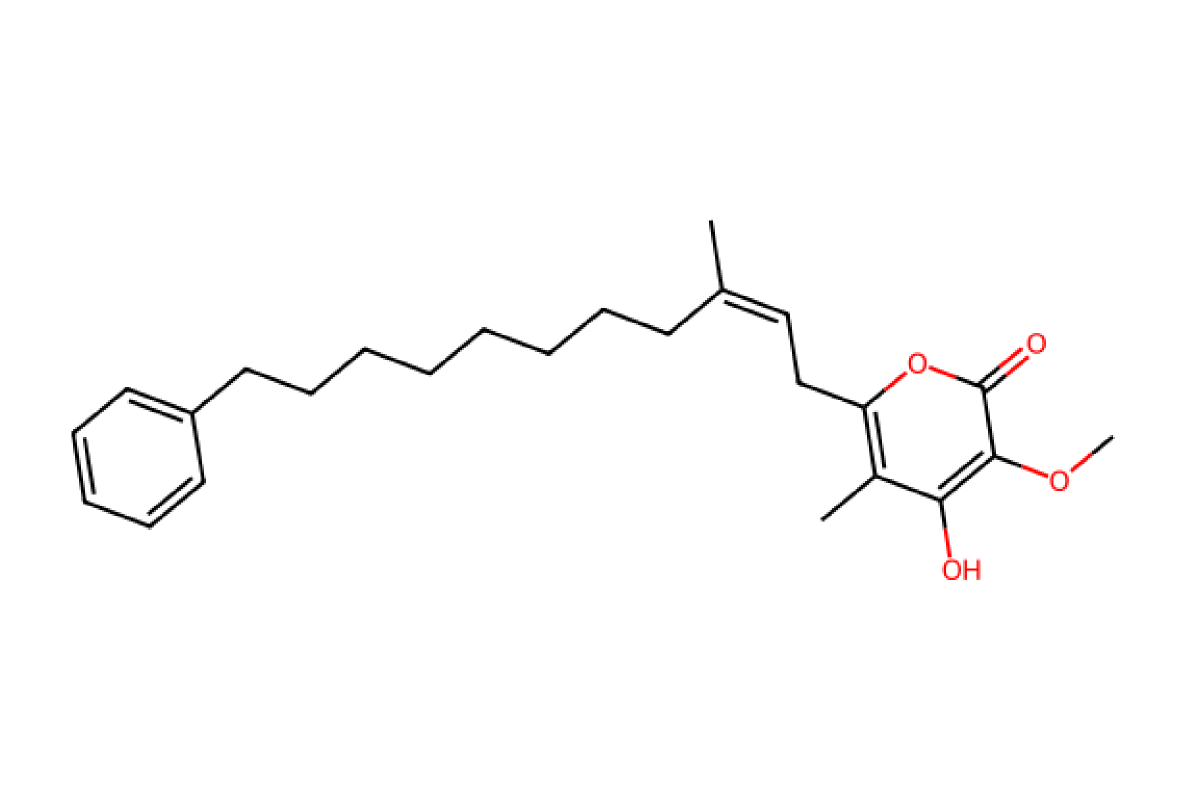} 
&
The molecule is a member of the class of \textcolor{matchcolor}{\textbf{2-pyranones}} that is \textcolor{matchcolor}{\textbf{2H-pyran-2-one}} substituted by a \textcolor{matchcolor}{\textbf{methoxy group at position 3}}, a \textcolor{matchcolor}{\textbf{hydroxy group at position 4}}, a methyl group at position 5 and a \textcolor{matchcolor}{\textbf{(2E)-3-methyl}}-11-phenylundec\textcolor{matchcolor}{\textbf{-2-en-1-yl group at position 6}}. Isolated from the marine sponge of the genus Plakortis, it exhibits cytotoxicity against human promyeloid leukemic HL-60 cells. It has a role as an \textcolor{matchcolor}{\textbf{antineoplastic agent}} and an animal metabolite. It is a member of \textcolor{matchcolor}{\textbf{2-pyranones}}, an \textcolor{matchcolor}{\textbf{ether}} and a \textcolor{matchcolor}{\textbf{polyketide}}.
&
The molecule is a member of the class of \textcolor{matchcolor}{\textbf{2-pyranones}} that is \textcolor{matchcolor}{\textbf{2H-pyran-2-one}} substituted by a \textcolor{matchcolor}{\textbf{methoxy group at position 3}}, a \textcolor{matchcolor}{\textbf{hydroxy group at position 4}}, \textcolor{diffcolor}{methyl groups at positions 5 and 6} and a \textcolor{matchcolor}{\textbf{(2E)-3-methyl}}-\textcolor{diffcolor}{9-phenylnon}\textcolor{matchcolor}{\textbf{-2-en-1-yl group at position 6}}. Isolated from the marine sponge of the genus Plakortis, it exhibits cytotoxicity against human promyeloid leukemic HL-60 cells. It has a role as a metabolite and an \textcolor{matchcolor}{\textbf{antineoplastic agent}}. It is a member of \textcolor{matchcolor}{\textbf{2-pyranones}}, an \textcolor{matchcolor}{\textbf{ether}} and a \textcolor{matchcolor}{\textbf{polyketide}}.\\
\midrule
\includegraphics[width=\linewidth, height=3cm, keepaspectratio]{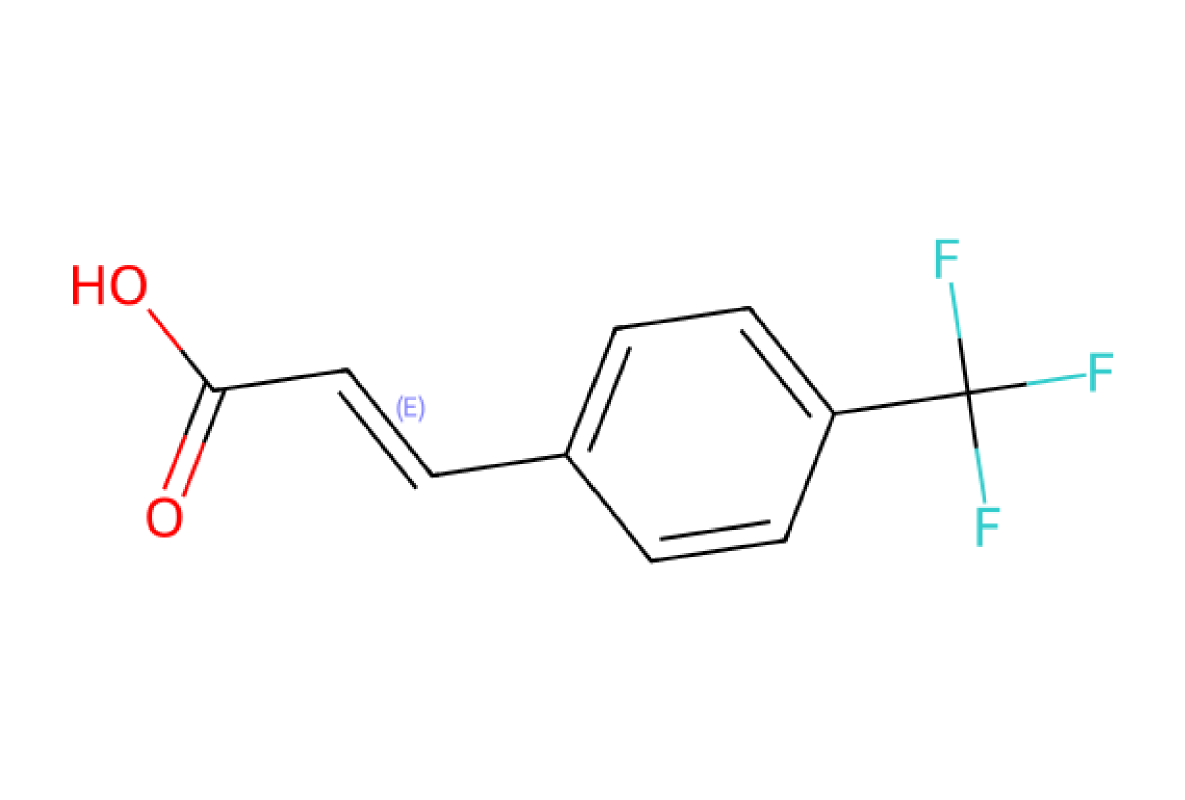} 
&
The molecule is a member of the class of \textcolor{matchcolor}{\textbf{(trifluoromethyl)benzenes}} consisting of \textcolor{matchcolor}{\textbf{trans-cinnamic acid}} having a \textcolor{matchcolor}{\textbf{trifluoromethyl substituent}} at the para-position. It is a member of \textcolor{matchcolor}{\textbf{cinnamic acids}} and a member of \textcolor{matchcolor}{\textbf{(trifluoromethyl)benzenes}}. It derives from a \textcolor{matchcolor}{\textbf{trans-cinnamic acid}}.
&
The molecule is a member of the class of \textcolor{matchcolor}{\textbf{(trifluoromethyl)benzenes}} consisting of \textcolor{matchcolor}{\textbf{trans-cinnamic acid}} having a \textcolor{matchcolor}{\textbf{trifluoromethyl substituent}} at the \textcolor{diffcolor}{meta-position}. It is a member of \textcolor{matchcolor}{\textbf{cinnamic acids}}, \textcolor{diffcolor}{a member of styrenes} and a member of \textcolor{matchcolor}{\textbf{(trifluoromethyl)benzenes}}. It derives from a \textcolor{matchcolor}{\textbf{trans-cinnamic acid}}. \\
\midrule
\includegraphics[width=\linewidth, height=3cm, keepaspectratio]{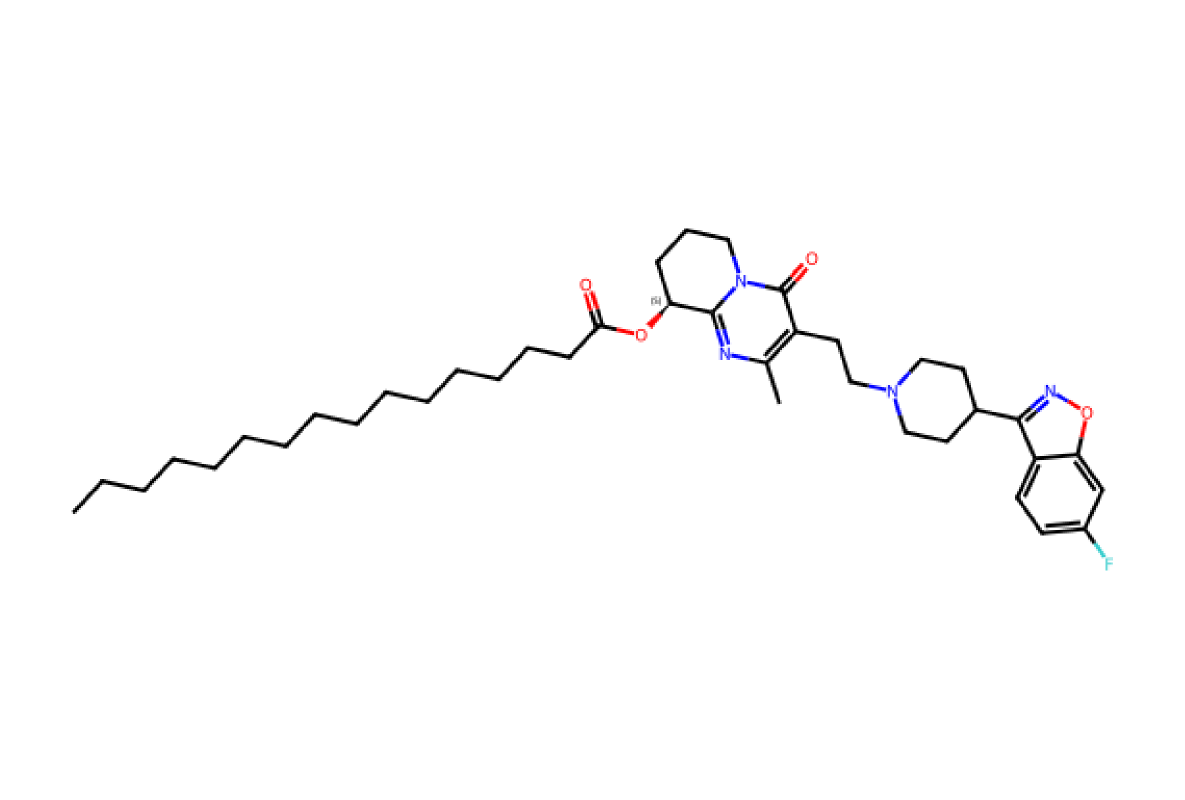} 
&
The molecule is a \textcolor{matchcolor}{\textbf{3-\{2-[4-(6-fluoro-1,2-benzoxazol-3-yl)piperidin-1-yl]ethyl\}-2-methyl-4-oxo-6,7,8,9-tetrahydropyrido[1,2-a]pyrimidin-9-yl hexadecanoate}} that is the \textcolor{matchcolor}{\textbf{(S)-enantiomer}} of \textcolor{matchcolor}{\textbf{paliperidone palmitate}}. It is an enantiomer of a (R)-paliperidone palmitate.
&
The molecule is a \textcolor{matchcolor}{\textbf{3-\{2-[4-(6-fluoro-1,2-benzoxazol-3-yl)piperidin-1-yl]ethyl\}-2-methyl-4-oxo-6,7,8,9-tetrahydropyrido[1,2-a]pyrimidin-9-yl hexadecanoate}} that is the \textcolor{diffcolor}{(R)-enantiomer} of \textcolor{matchcolor}{\textbf{paliperidone palmitate}}. It is an enantiomer of a \textcolor{diffcolor}{(S)-paliperidone palmitate}. \\
\midrule
\includegraphics[width=\linewidth, height=3cm, keepaspectratio]{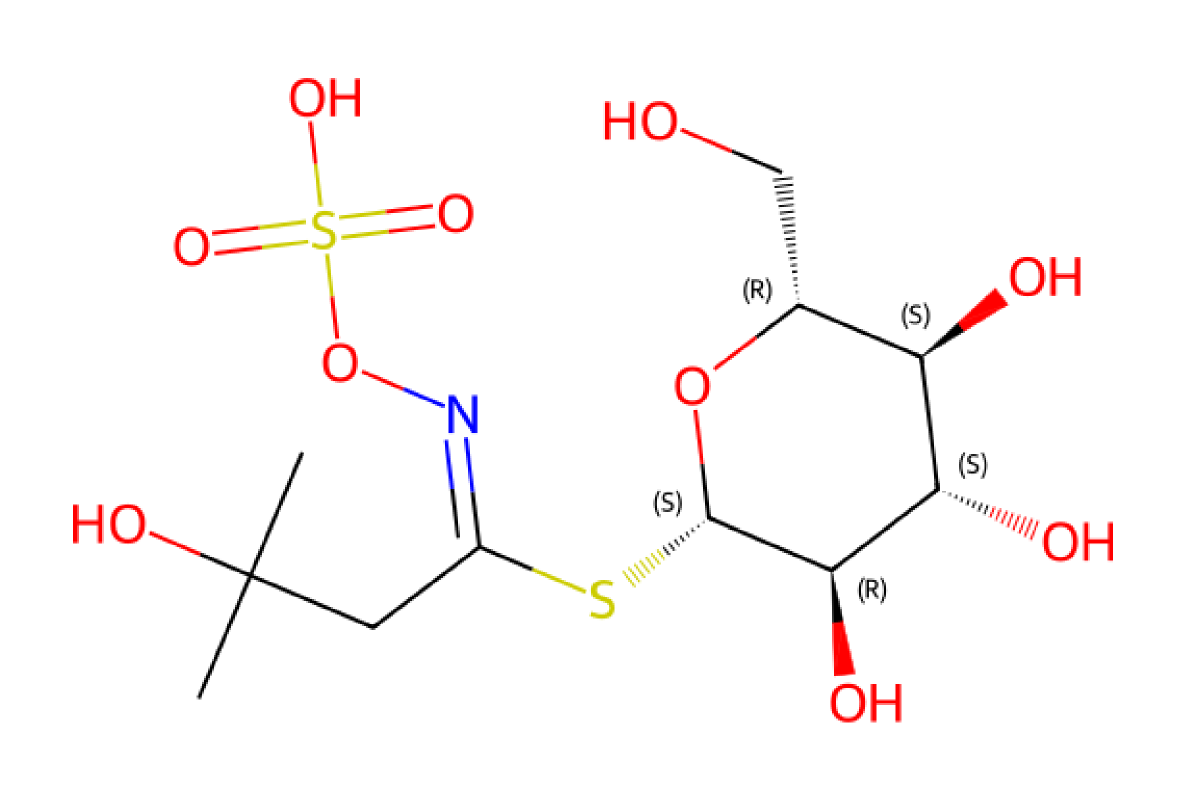} 
&
The molecule is a \textcolor{matchcolor}{\textbf{hydroxy-alkylglucosinolic acid}} that consists of \textcolor{matchcolor}{\textbf{1-thio-beta-D-glucopyranose}} attached to a \textcolor{matchcolor}{\textbf{3-hydroxy-3-methyl-N-(sulfooxy)}}butanimidoyl group at the \textcolor{matchcolor}{\textbf{anomeric sulfur}}. It derives from an \textcolor{matchcolor}{\textbf{isobutylglucosinolic acid}}. It is a conjugate acid of a glucoconringiin(1-).
&
The molecule is a \textcolor{matchcolor}{\textbf{hydroxy-alkylglucosinolic acid}} that consists of \textcolor{matchcolor}{\textbf{1-thio-beta-D-glucopyranose}} attached to a \textcolor{matchcolor}{\textbf{3-hydroxy-3-methyl-N-(sulfooxy)}}\textcolor{diffcolor}{pentanimidoyl} group at the \textcolor{matchcolor}{\textbf{anomeric sulfur}}. It derives from an \textcolor{matchcolor}{\textbf{isobutylglucosinolic acid}}. It is a conjugate acid of a \textcolor{diffcolor}{glucocleomin(1-)}. \\
\bottomrule
\end{tabular}
\caption{Comparison of ground truth and MoRA descriptions. Matching chemistry-related parts are highlighted in {\color{matchcolor}\textbf{blue}}, while different details are marked in {\color{diffcolor}red}.}
\label{tab:desc_comparison}
\end{table*}
\newpage
\bibliography{aaai2026}
\end{document}